\documentclass{article}
\usepackage{multirow}
\usepackage{microtype}
\usepackage{graphicx}
\usepackage{subfigure}
\usepackage{booktabs} 
\usepackage{bbm}
\usepackage{lipsum}
\usepackage{csquotes}
\usepackage{xcolor}

\usepackage{hyperref}

\usepackage[accepted]{icml2024}

\usepackage{amsmath}
\usepackage{amssymb}
\usepackage{mathtools}
\usepackage{amsthm}
\usepackage{bm}

\usepackage[capitalize,noabbrev]{cleveref}

\theoremstyle{plain}
\newtheorem{theorem}{Theorem}[section]

\theoremstyle{definition}
\newtheorem{definition}[theorem]{Definition}
\newtheorem{assumption}[theorem]{Assumption}
\theoremstyle{remark}

\usepackage[textsize=tiny]{todonotes}
\usepackage{breqn}

\newcommand{\methodname}{{\tt{FedCal}}}

\icmltitlerunning{\methodname{}: Model Calibration in Federated Learning via Aggregated Parameterized Scaler}

\begin{document}

\twocolumn[
\icmltitle{\methodname{}: Achieving Local and Global Calibration in Federated Learning via Aggregated Parameterized Scaler}




\begin{icmlauthorlist}
\icmlauthor{Hongyi Peng}{NTU}
\icmlauthor{Han Yu}{NTU}
\icmlauthor{Xiaoli Tang}{NTU}
\icmlauthor{Xiaoxiao Li}{UBC,Vector}

\end{icmlauthorlist}

\icmlaffiliation{NTU}{College of Computing and Data Science, Nanyang Technological University, Singapore.}
\icmlaffiliation{UBC}{Department of Electrical and Computer Engineering, The University of British Columbia, Vancouver, BC, Canada.}
\icmlaffiliation{Vector}{Vector Institute, Canada}

\icmlcorrespondingauthor{Han Yu}{ han.yu@ntu.edu.sg}

\icmlkeywords{Federated Learning, Model Calibration, Data Heterogeneity}

\vskip 0.3in
]



\printAffiliationsAndNotice{} 
\begin{abstract}

Federated learning (FL) enables collaborative machine learning across distributed data owners. However, this
approach poses a significant challenge for model calibration due to data heterogeneity. While prior work focused on improving accuracy for non-iid data, calibration remains under-explored. This study reveals existing FL aggregation approaches lead to sub-optimal calibration, and theoretical analysis shows despite constraining variance in clients' label distributions, global calibration error is still asymptotically lower bounded. To address this, we propose a novel \underline{Fed}erated \underline{Cal}ibration (\methodname{}) approach, emphasizing both local and global calibration. It leverages client-specific scalers for local calibration to effectively correct output misalignment without sacrificing prediction accuracy. These scalers are then aggregated via weight averaging to generate a global scaler, minimizing the global calibration error.  Extensive experiments demonstrate that \methodname{} significantly outperforms the best-performing baseline, reducing global calibration error by 47.66\% on average.

\end{abstract}

\section{Introduction}
\label{introduction}
Federated learning (FL) \citep{McMahan_Moore_Ramage_Hampson_Arcas_2017,liu2024federated} has emerged as a novel distributed machine learning paradigm, enabling clients to train models collaboratively. A fundamental challenge in this domain is the data distribution heterogeneity among FL clients (i.e., the data non-IIDness issue). Numerous studies~\citep{kairouz2021advances, li2020federated, zhao2018federated} have demonstrated that this issue can adversely affect the accuracy and convergence of FL models. In response, extensive research efforts on personalized federated learning (PFL) \citep{tan2023towards, li2020federated, karimireddy2020scaffold, wang2020tackling, reddi2020adaptive} have been directed towards mitigating this issue.

However, an aspect that has been largely overlooked in the current PFL literature is \textit{reliability} \citep{lu2021distribution, pmlr-v202-plassier23a}. As FL systems become increasingly embedded in high-stake decision-making processes, the importance of reliability, especially in mission-critical applications such as healthcare \citep{rieke2020future, dayan2021federated, sheller2020federated}, finance \citep{long2020federated, dash2022federated} and autonomous driving systems \citep{li2021privacy, du2020federated}, cannot be overstated. 

Besides prediction accuracy, the reliability of an FL model also hinges on accurate uncertainty estimation, commonly referred to as \textit{confidence} \citep{Ovadia_Fertig_Ren_Nado_Sculley_Nowozin_Dillon_Lakshminarayanan_Snoek_2019, yu2022robust, Guo_Pleiss_Sun_Weinberger_2017}. Consider a model for cancer diagnosis. It predicts that a patient has cancer with a confidence level of 0.1. This confidence score carries significant implications. For instance, a patient predicted with a 0.1 confidence might receive a more cautious treatment recommendation. In a well-calibrated model, if 10 patients are assigned a confidence of 0.1 for having cancer, approximately one of them should genuinely be diagnosed correctly with the disease. This process of aligning the model prediction confidence with the actual observed frequency of an event is referred to as \textit{model calibration} \citep{guo2017calibration}. 

\begin{figure}[!t]
\vskip 0.2in
\begin{center}
\centerline{\includegraphics[width=\linewidth]{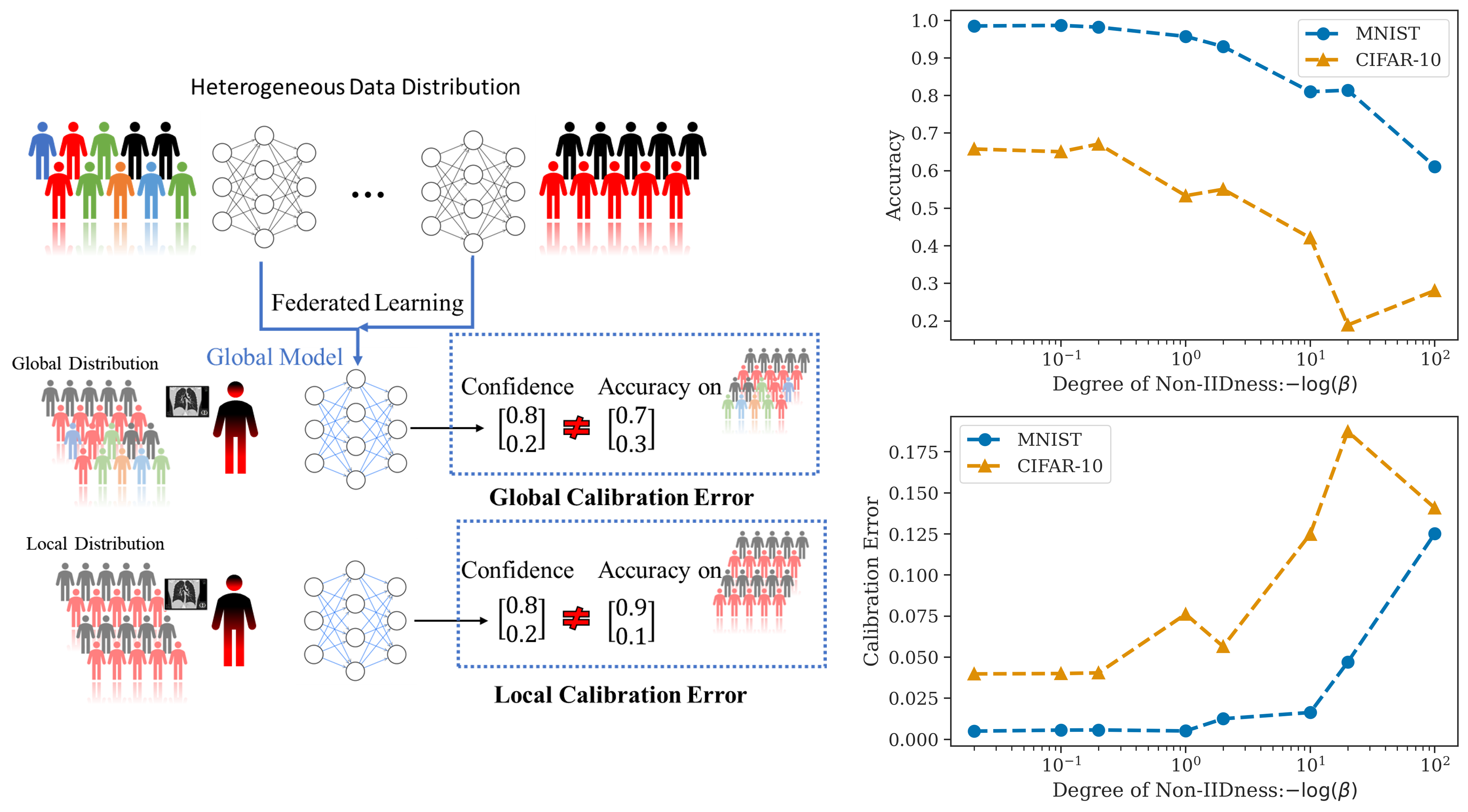}}

\caption{
[Left] Impacts of Data Distribution Discrepancies on Model Calibration in Federated Learning. The presence of non-IID data across local nodes contributes to miscalibration issues in the aggregated model, influencing both local and global datasets. [Right] Impact of data heterogeneity on the accuracy and reliability of FL models. As the degree of non-IIDness (quantified by the Dirichlet distribution parameter $\beta$) increases, both accuracy and reliability of FL models trained on MNIST (MLP, 10 clients) and CIFAR-10 (ResNet-14, 10 clients) using FedAvg \citep{McMahan_Moore_Ramage_Hampson_Arcas_2017} deteriorate.} 
\label{fig:central vs fedavg comparison}
\end{center}
\vskip -0.2in
\end{figure}

Model calibration is of critical importance to producing reliable machine learning models. While the topic has been extensively examined under centralized learning settings, it has been largely overlooked under FL settings. As highlighted in Figure \ref{fig:central vs fedavg comparison}, the prevailing FedAvg-based \cite{McMahan_Moore_Ramage_Hampson_Arcas_2017} FL model aggregation approaches produce poorly calibrated models in the presence of data heterogeneity. Recognizing this gap, we delve into the unique challenges posed by FL, and contend that existing centralized calibration techniques are not directly applicable in the face of the distributed and data heterogeneous nature of FL.

Model calibration in FL settings faces two major challenges:
\begin{enumerate}
    \item \textbf{Need for Local and Global Calibration.} Our empirical observations reveal a dual challenge for calibration in data heterogeneous FL. Firstly, data heterogeneity negatively impacts the overall calibration of the global model. Secondly, the variance of calibration errors across clients increases with the degree of data heterogeneity. This necessitates a two-pronged approach: a personalized calibration step adjusting the model to individual client's data distribution, and a robust global calibration step for improved generalizability.
    \item \textbf{Lack of Global Validation Datasets.} Given the distributed nature of FL, it is often impractical to have a comprehensive global validation set for global calibration due to privacy concerns and the challenges of data collection and maintenance. 
\end{enumerate}


To bridge this gap, our goal is to achieve both local and global model calibration via FL aggregation without relying on the existence of a global validation dataset.
To achieve this goal, we propose a novel \underline{Fed}erated \underline{Cal}ibration (\methodname{}) approach. It involves training post-hoc scalers on local datasets for local calibration and subsequently aggregating them to achieve global calibration. Notably, the scalers in \methodname{} can be aggregated simply by averaging their parameters.
As a post-hoc calibration method, \methodname{} can be easily integrated with existing FL methods.

Theoretical analysis shows that despite constraining the variance in clients' label distributions, the global model calibration error still asymptotically decreases.
Extensive experimental evaluation based on four benchmark datasets reveals that \methodname{} significantly surpasses five state-of-the-art methods in both local and global calibration, regardless of the presence of global validation sets. Specifically, it reduces the global model calibration error by 47.66\% on average compared to the best-performing baseline. Furthermore, we observe that ensembling global and local models can further enhance prediction accuracy. 



\section{Related Work}
\label{related works}


\paragraph{Calibration} is an important research topic in centralized ML. A vast body of literature exists on the calibration of finely-tuned ML models. Notable methods include histogram binning \citep{zadrozny2001obtaining}, isotonic regression \citep{zadrozny2002transforming}, conformal prediction \citep{vovk2005algorithmic}, Platt scaling \citep{platt1999probabilistic}, and temperature scaling \citep{guo2017calibration}. These techniques generally rely on the existence of a validation set for the post-processing of model predictions.

Recent research has started to focus on enhancing calibration in deep learning models. These include strategies such as augmentation-based training \citep{thulasidasan2019mixup}, calibration in neural machine translation \citep{kumar2019calibration}, neural stochastic differential equations \citep{kong2020sde}, self-supervised learning \citep{hendrycks2019using} ensemble methods \citep{lakshminarayanan2017simple}, and even providing statistical assurance for calibration in black-box models \citep{angelopoulos2021learn}. Nevertheless,  research addressing model calibration in FL settings remains limited.

\paragraph{Calibration in FL Settings}
While the importance of calibration in FL is being recognized, existing approaches primarily focus on performance improvement without considering reliability.
\citet{fl-calibration-luo2021no} demonstrated that classifier calibration significantly boosts FL model performance, albeit focusing on adjusting weights on IID data, a process different from aligning confidence with observed frequency. \citet{fl-calibration-zhang2022federated} introduced FedLC to enhance model accuracy through a calibration-inspired cross-entropy loss. However, calibration in FedLC is defined as the error rate per class, which is different from our definition. \citet{fl-calibration-achituve2021personalized} proposed pFedGP, a personalized approach that, although not designed for FL calibration, achieves some level of empirical calibration efficacy. But, its specialized nature hinders wider applicability. Closely related to our work is MD-TS \cite{yu2022robust}. It employs domain-specific temperature scaling and a predictive linear regression model. Nevertheless, it relies on the existence of global validation sets, which might be difficult to prepare in practice and risk privacy leakage.

Orthogonal to our work, the field of PFL tailors models to client-specific needs \citep{tan2023towards, arivazhagan2019federated, deng2020adaptive}. While PFL addresses the issue of non-IID client data distributions, our focus lies in federated model calibration. Notably, \methodname{} can be integrated with existing PFL frameworks.

\section{Preliminaries}

\newcommand{\vect}[1]{\bm{#1}}
\newcommand{\propk}{f_k(\vect{x}_i)}
\newcommand{\topProp}{\hat{f}(\vect{x}_i)}
\newcommand{\prob}{\text{I\kern-0.15em P}}
\newcommand{\expectation}{\mathop{\mathbb{E}}}
\newcommand{\indicator}{\mathbbm{1}}
\newcommand{\pred}{\hat{y}(\vect{x}_i)}

\subsection{Basics of Model Calibration}
In a $K$-classification task with $\mathcal{Y} \equiv \{1, ..., K\}$, we aim to train a model
$\vect{\theta}: \mathcal{X} \mapsto \mathbb{R}^K$ that predicts the labels $y\in\mathcal{Y}$ corresponding to the input $\vect{x}\in\mathcal{X}$. Here, $\vect{\theta}(\vect{x}_i)$ outputs the score, a quantity that represents a proper probabilistic estimation often obtained by applying an additional sigmoid layer $\sigma$ as $f(\vect{x}_i) = \sigma(\vect{\theta}(\vect{x}_i))$, where $f: \mathcal{X} \mapsto \Delta^{K-1}$. $\Delta^{K-1}$ denotes a $(K-1)$ simplex such that for any class $k$, $f_k(\vect{x}_i) \in [0, 1]$ and $\sum_{k\in \mathcal{Y}} f_k(\vect{x}_i) = 1$.

Calibration \cite{guo2017calibration}, in this framework, refers to the extent to which these predicted probabilities $\propk$ reflect the true conditional probability $p(y=k|\vect{x})$. While calibrating predictions for every class is an ideal goal, it can be challenging and impractical in real applications \citep{kumar2019verified}. Often, the more achievable task of top-label calibration is prioritized, which focuses on ensuring the predicted probability of the most likely class aligns with its true probability. For a comprehensive overview of different calibration types and their nuances, please refer to \citep{kumar2019verified, zhao2021calibrating}. 

We denote the predicted label $\topProp$ by a model as $\topProp \equiv \max \propk$ and $\hat{y}(\vect{x}) = \arg\max_{k\in \mathcal{Y}}\propk$. Then, the top-level calibration error can be defined as: 
\begin{definition} (Calibration Error). The calibration error $\text{CE}(f)$ of $f$ is given by:
\begin{equation}
\text{CE}(f) = (\expectation[(\prob(y=\pred|\topProp) - \topProp)^2])^{\frac{1}{2}}.
\end{equation}\label{eq:ce_def}
\end{definition}
The lower the CE value, the better the calibration. A perfectly calibrated model achieves a CE value of 0. A widely adopted empirical measure of CE is the Expected Calibration Error (ECE) \citep{naeini2015obtaining}. ECE measures calibration by averaging the difference between the predicted probabilities and the actual accuracy within each confidence bin. In our work, ECE is adopted as the metric for reporting calibration errors.

\begin{definition} (Expected Calibration Error). Giving a partition $c_m$ of the unit interval $[0, 1]$ and the buckets $B_m = \{ i: c_{m-1} < \topProp \leq c_m\}$, ECE is defined as:
\begin{equation}
    \text{ECE} = \sum_{m=1}^{M} \frac{|B_m|}{N}|\text{conf}_m - \text{acc}_m|,
\end{equation}
where $\text{acc}_m = \frac{1}{|B_m|}\sum_{i\in B_m}\indicator(\pred = y)$ and $\text{conf}_m = \frac{1}{|B_m|}\sum_{i\in B_m}\propk$.
\end{definition}

\subsection{Calibration by Scaling}
Modern deep models tend to achieve poor calibration \cite{guo2017calibration}. To address this issue, researchers have explored two main approaches: 1) post-hoc calibration, and 2) architecture/training modification for inherently calibrated models \cite{wu2021should}. We focus on the post-hoc calibration approach due to its flexibility and ease of integration with existing training paradigms. Post-hoc mapping methods involve employing a scaling function $\phi$ trained on an auxiliary dataset to adjust $\propk$. Here, we review a simple yet effective calibration method, Temperature scaling \cite{guo2017calibration}, which resales model outputs using a singular temperature parameter $T$ as:
\begin{equation}
    \phi(T) \circ f(\vect{x}_i)\equiv \sigma\left(\frac{\vect{\theta}(\vect{x}_i)}{T}\right).
\end{equation}
The optimal temperature value, $T$, is determined by minimizing the negative log-likelihood (NLL) on a validation set $\mathcal{D}_{val}$ as:
\begin{equation}
\begin{aligned}
    \min_{T} & \text{NLL}\left(y, \sigma\left(\frac{\vect{\theta}(\vect{x}_i)}{T}\right)\right) \\
    &= -\sum_{i=1}^{|\mathcal{D}_{val}|}\sum_{j=1}^K \indicator(y_i=j) \cdot \log \sigma_j\left(\frac{\vect{\theta}(\vect{x}_i)}{T}\right).
\end{aligned}\label{eq:get_scaler}
\end{equation}


\section{Theoretical Basis of \methodname{}}

\begin{figure}[!t]
\begin{center}
\centerline{\includegraphics[width=\linewidth]{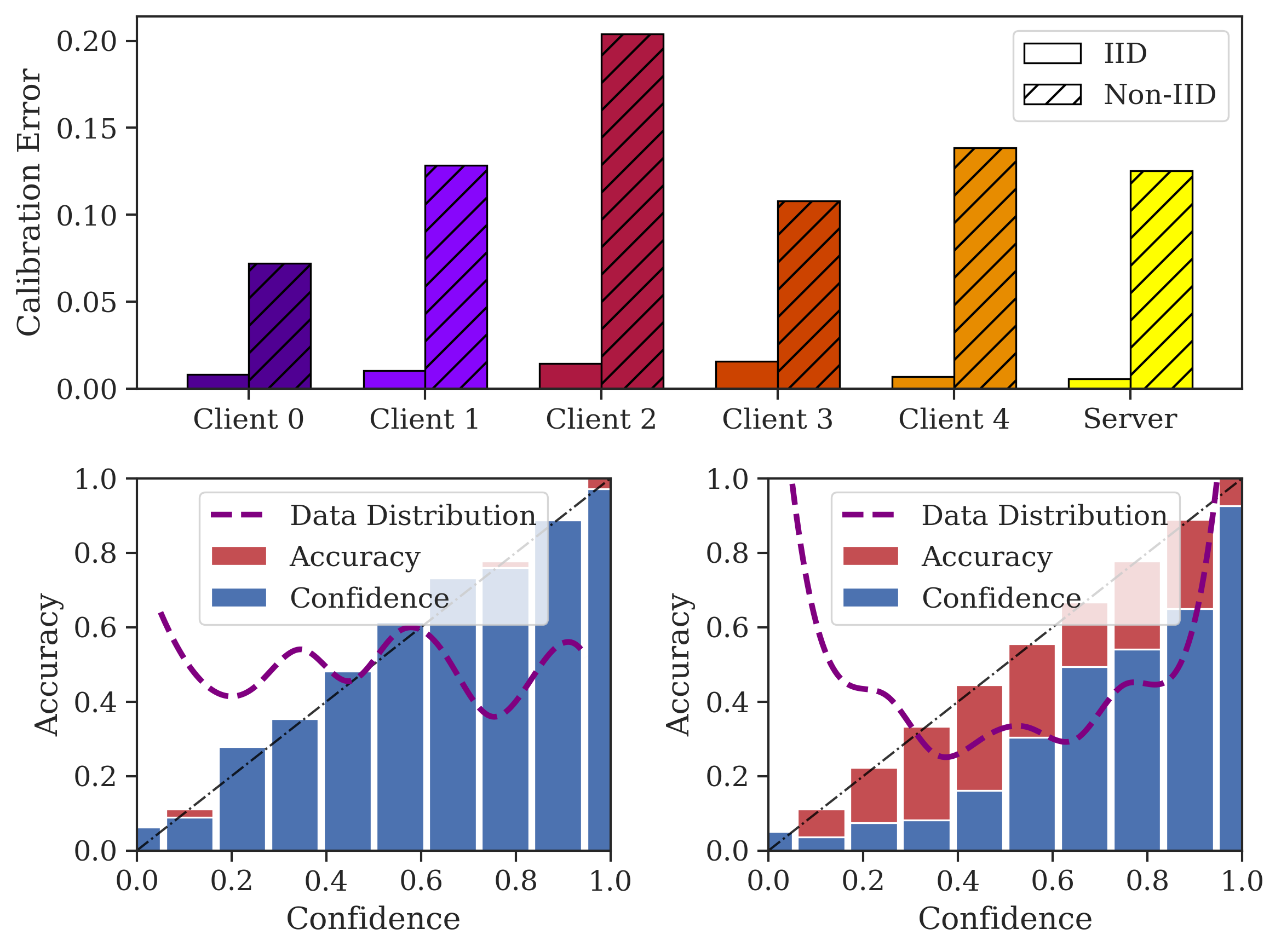}}
\caption{Impact of Non-IID Data Distribution on Client and Server Calibration in FL. The top plot shows the calibration error of five clients trained on the MNIST dataset with a Multilayer Perceptron (MLP) model under IID and non-IID distributions. We observe that the non-IID client exhibits significantly higher calibration error compared to the IID clients, and that calibration error can vary significantly across clients and even at the server due to the skewed data distribution. The two bottom plots depict the model reliability for a single client trained under IID and non-IID conditions. The purple dashed line represents the normalized class density of a client, and the gray dashed line represents perfect calibration (i.e., confidence aligns exactly with accuracy). In the IID case (left plot), the model is well-calibrated, with the confidence closely matching the accuracy throughout the range. In contrast, the non-IID case (right plot) reveals severe under-confidence.}
\label{fig: local and global calibration}
\end{center}
\end{figure}

\subsection{The Need for Local and Global Calibration}
Calibrating FL models introduces distinct challenges due to the inherent non-IID nature of data across clients. Such heterogeneous data distributions significantly degrade the calibration performance of models evaluated on both local and global datasets (Figure \ref{fig: local and global calibration}). Moreover, the privacy requirements inherent to FL often preclude access to a centralized auxiliary validation dataset, rendering traditional calibration methods inapplicable. Figure \ref{fig: local and global calibration} also highlights the discrepancies between calibration performance on clients' local datasets and the aggregated global dataset. This observation compels us to re-evaluate traditional notions of calibration within the context of FL. We posit that both local and global calibration are essential for building reliable FL models.

Consider the illustrative hospital example from the introduction, where treatment decisions hinge on the model's prediction and the associated confidence, interpreted as cancer probability. The decision minimizes the combined surgical and conservative treatment risks: \( \textit{risk}_{\text{surgery}} \times \prob(\text{cancer}) + \textit{risk}_{\text{conservative treatment}} \times \prob(\text{benign tumor}) \). As posited by \citet{zhao2021calibrating}, calibration directly impacts decision-making efficacy. Hospital A's specialized model, potentially well-calibrated for its specific patient population, might predict a higher cancer probability, leading to a more aggressive treatment approach. Conversely, Hospital B's general model, trained on a wider range of patients with different diseases, might predict a lower probability, favoring a more conservative approach. While local expertise and specialization hold value, neglecting global calibration across the FL network poses significant risks.

Maintaining consistent global calibration ensures that decisions are guided by comparable risk assessments across all participating clients. This, in turn, minimizes the potential for inequitable recommendations, especially as new clients with potentially divergent data distributions join FL. Here, we formalize the notions of local and global calibration.

\begin{definition}(Local calibration error and global calibration error).
Consider an FL system comprising $C$ clients, each possessing a local dataset $\mathcal{D}_c = \{\vect{x}_i, y_i\}_{i=1}^{N_c}$ drawn from a client-specific distribution $\prob_c(\vect{x}, y)$. The local calibration error $\text{CE}_c(f)$ is:
\begin{equation}
\begin{aligned}
    \text{CE}_c(f) = \Bigg( \expectation_{\prob_c(\vect{x}, y))}[(\prob(y=\pred|\topProp) - \topProp)^2]\Bigg) ^{\frac{1}{2}}.
\end{aligned}
\end{equation}
Similarly, the global calibration error $\overline{\text{CE}}(f)$, conceptualized as the expected calibration error over unseen data sampled from a client-agnostic distribution $\prob(\vect{x}, y)$, is: 
\begin{equation}
\overline{\text{CE}}(f) =  \Bigg(\expectation_{\prob(\vect{x}, y)}[(\prob(y=\pred|\topProp) - \topProp)^2]\Bigg)^{\frac{1}{2}}.
\end{equation}\label{eq:global_calibration_error}
\end{definition}

Empirically, local calibration error can be approximated by evaluating ECE on the local dataset. Since the global data distribution is not directly observable, it is presumed that samples $(\vect{x}_i, y_i)$ in global dataset follows the same distribution of the pooled dataset $\cup_{c=1}^{C}\mathcal{D}_c$ and the global calibration error can be assessed using a reserved test set, sampled from the pooled dataset.

\subsection{Label Skew Leads to Poor Calibration}
In this paper, we focus on one of the most representative settings of non-IIDness, which is \textit{label skew} \citep{luo2019real, lyu2022privacy, zhang2022federated}. Note that, while we focus on label skew, various forms of non-IID data exist (refer to \cite{hsieh2020non} for more details).

\begin{definition} (Label Skew). The label distribution across the clients is skewed. If the local distribution is rewritten as $\prob_c(\vect{x}, y) = \prob_c(\vect{x}|y)\prob_c(y)$, label skew means, for any two clients $c_1$ and $c_2$, we have: (1) $\prob_{c1}(y)\neq\prob_{c2}(y)$ if $c_1 \neq c_2$; and (2) $\prob_{c1}(\vect{x}|y) = \prob_{c2}(\vect{x}|y)$.
\end{definition}\label{def:label_skew}

We denote the empirical risk minimizer over the global distribution as $f^*(\vect{x}) \coloneqq \arg_{f} \min \xi(y, f(\vect{x}))$, where $\xi$ is the cross entropy loss. The global model obtained after $r$ rounds by averaging local updates is denoted as $f^r(\vect{x})$. Affected by the heterogeneous data, the local objectives of different clients are generally not identical and might not share the same risk minimizer. Consequently, even as all clients start from the same global model, local updates will steer the model towards the minima of local objectives (known as client drift \citep{charles2021convergence}). This divergence implies that averaging local updates via FedAvg results in a model that is different from the global minimizer, i.e., $f^*(\vect{x}) \neq f^r(\vect{x})$ \citep{wang2020tackling, karimireddy2020scaffold}. Expanding upon this fundamental understanding, we posit that, given a bounded distribution divergence between local and global distributions, client drift inherently imposes an asymptotic lower bound on the global calibration error.

\begin{assumption}(Bounded discrepancy between local and global label distribution). 
We assume that the discrepancy between the data distributions at each client and the global distribution, measured by the Kullback–Leibler divergence $\text{D}_{KL}$, is bounded.
Specifically, the maximum divergence between the local data distribution $\prob_c(y)$ and the global distribution $\prob(y)$ does not exceed a value $G$. Mathematically, this is expressed as:
\begin{equation}
\sup_{c \in \{1, \cdots, C\}} \text{D}_{KL}(\prob_c(y) \parallel \prob(y)) \leq G,
\end{equation} 
where $C$ is the total number of clients.
\end{assumption}\label{ass:distribution_divergence}

For clarity and ease of understanding, we concentrate the discussion on the binary classification scenario. This does not reduce the generality of our findings as the problem of calibrating multi-class models can be effectively reduced to the binary case by letting the model output a probability corresponding to its top prediction, and the label represents whether the prediction is correct or not \citep{kumar2019verified}.

\begin{theorem}(Lower bound of global calibration error). Consider the scenario where the discrepancy between the local and global label distributions is bounded by $G$, as stated in Assumption \ref{ass:distribution_divergence}. Let $R$ represent the number of FL communication rounds involving more than two clients. Jointly considering the established assumptions detailed in Appendix \ref{proof}, under these conditions, there exists a $\mu$-convex objective function for which the resulting global model $f^R(\vect{x})$ after $R$ rounds has a calibration error asymptotically bounded by:
\begin{equation}
    \overline{\text{CE}}(f^R) \geq \Omega\bigg(\frac{\sqrt{\frac{1}{2}G}}{\mu R^2}\bigg).
\end{equation}\label{th:global_calibration error}
\end{theorem}

The formal proof of this theorem is presented in Appendix \ref{proof}. Here, we provide an illustrative sketch. Firstly, we establish that the global risk minimizer $f^*(\vect{x})$ also minimizes the calibration error (i.e., $\overline{\text{CE}}(f^*(\vect{x}))=0$). Consistent with the premises set forth in \citep{zhang2022federated, wang2021addressing}, we posit that the features extracted from the same class (inputs to the final layer) exhibit a high degree of similarity, a notion empirically observed and verified in \citep{wang2021addressing}. Based on this, we analyse how the bounded discrepancy between label distributions propagates to bounded gradient dissimilarity using Pinsker's Inequality. Lastly, based on \citet{karimireddy2020scaffold} which relates the dissimilarity of gradient to the difference between risk minimizer $[f^R(\vect{x}) - f^*(\vect{x})]$, we draw the proof to its conclusion.

Theorem \ref{th:global_calibration error} elucidates that despite constraining the variance in clients' label distributions, the global model calibration error can still asymptotically lower bounded, which corroborates our empirical findings depicted in Figure \ref{fig: local and global calibration} and Figure \ref{fig:central vs fedavg comparison}.
While prevailing calibration techniques (e.g., temperature scaling) offer simplicity and effectiveness, they require access to a global validation dataset. This is not only impractical in many real-world scenarios, but also raises significant privacy concerns.

\methodname{} is a novel paradigm distinctly orthogonal to the conventional accuracy-centric FL approach. Formally, it is a task that aims to minimize both $\text{CE}_i$ and $\overline{\text{CE}}$ without access to $\mathcal{D}_{val}$ sampled from $\mathcal{D}$.

\section{The Proposed \methodname{} Approach}\label{scaler_aggregation}

While local calibration is relatively easy as the FL client has direct access to its local data making established calibration methods like temperature scaling applicable, achieving global calibration presents a significant technical challenge. To this end, we frame the problem of \methodname{} as aggregating local scalers into a global scaler, formulated as $\overline{\phi} = \text{Agg}({\phi_c}_{i=1}^C)$. The success of $\overline{\phi}$ hinges on two key aspects: 1) the scaler architecture, and 2) the aggregation strategy. The following properties are desirable for these two components:
\begin{enumerate}
    \item The scaler must possess robust generalization capabilities to handle potential discrepancies between local and global data distributions.\label{prop1} 
    \item The processes of scaling and calibration should maintain model accuracy.\label{prop4}
    \item The scaler should also be ``aggregatable", meaning that the aggregated version should perform well not only on a global scale, but also on the local scale.\label{prop2}
    \item The scaler aggregation strategy should not require direct access to local data distributions.\label{prop3}

\end{enumerate}

\subsection{Scaler Architecture Design}
To achieve Property \ref{prop1}, \methodname{} is equipped with a multi-layer perceptron (MLP) with substantially more parameters than traditional minimal-parameter methods to enhance its generalization capability. The MLP processes the original output logits $\vect{\theta}(x)$, transforming them into accurate probabilities, particularly under conditions of significant local and global data distribution discrepancies.

\begin{figure}[!b]
\begin{center}
\centerline{\includegraphics[width=\linewidth]{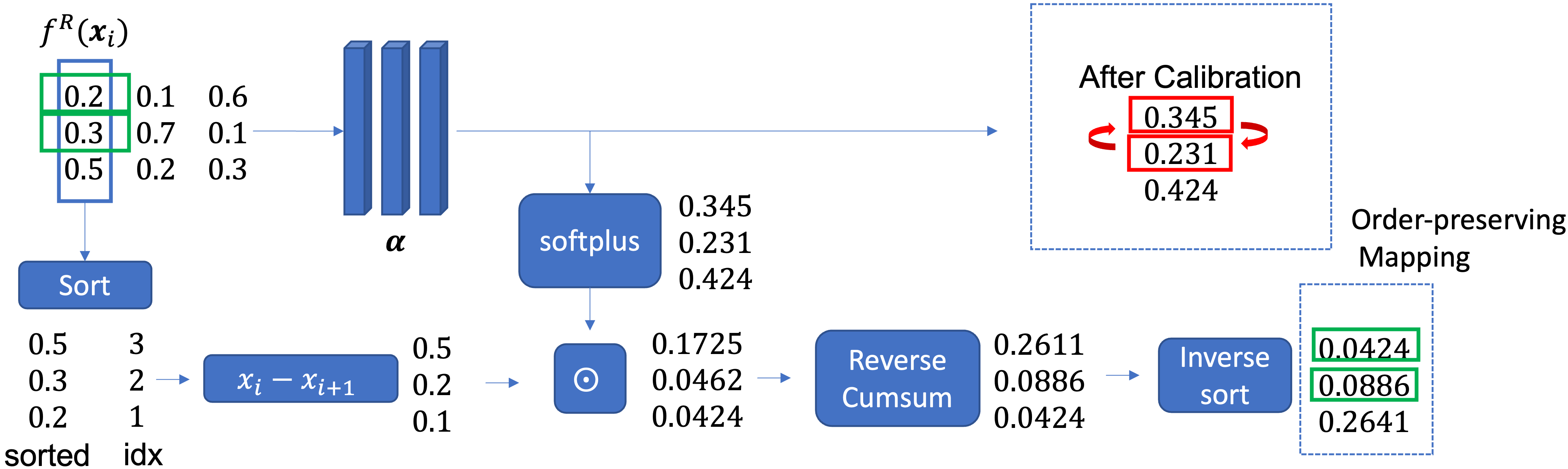}}
\caption{Impact of the Order-Preserving Network. Without order preservation, $\alpha$ can represent arbitrary mappings, potentially altering the predicted class ordering (highlighted in red). The working principles of order-preserving networks are in Appendix \ref{example_op}.}
\label{fig: order-preserving network}
\end{center}
\end{figure}

However, a trade-off exists between Property \ref{prop1} and Property \ref{prop4}. A more complex model, while capable of learning intricate mappings, risks overfitting the validation datasets. This can alter the original order of the model logits, leading to a decrease in top-k accuracy (Figure \ref{fig: order-preserving network}).
Here order-preserving means that for any two classes 
$j$ and $k$, if $f^R_j(\vect{x}_i) > f^R_k(\vect{x}_i)$, the scaled outputs must follow the same order (i.e., $\phi(f^R_j(\vect{x}_i)) > \phi(f^R_j(\vect{x}_i))$).
To achieve order preservation, we incorporate the order-preserving technique from \cite{rahimi2020intra} into the scaler design.
Formally, the local scaler $\phi_c$ parameterized by an MLP $\vect{\pi}_c$ remaps the model output as:
\begin{equation}
    \phi_c(\vect{\phi}) \circ f(\vect{x}_i)\equiv \phi_c(\vect{\theta}(\vect{x}_i); \vect{\pi}_c).
\end{equation}
$\vect{\pi}_c$ is optimized by minimizing the NLL loss on a local dataset $\mathcal{D}_c$ similar to Eq. \eqref{eq:get_scaler}. 
Although training is required, the scaler can be effectively implemented in a post-hoc manner. Specifically, the local scaler undergoes training only after the FL model completes its local training phase. 

\subsection{Aggregation Strategy Design}
One advantage of our MLP-based scaler design is its inherent permutation symmetries. These symmetries facilitate the aggregation of two distinct MLPs through linear mode connectivity \cite{ainsworth2022git, entezari2021role, nguyen2021solutions}. Linear mode connectivity implies that for two MLPs, $\vect{\pi}_i$ and $\vect{\pi}_j$, after aligning one of them (e.g., $\vect{\pi}_i$) through a weights permutation $\mathbf{M}(\vect{\pi}_i)$, they can be linearly combined in their parameter space as
$\vect{\pi}^* = \lambda \vect{\pi}_i + (1-\lambda) \mathbf{M}(\vect{\pi}_j)$. The loss function of the new parameters $\mathcal{L}(\vect{\pi}^*)$ closely approximate a weighted sum of the individual losses as $\mathcal{L}(\vect{\pi}^*) \approx \lambda \mathcal{L}(\vect{\pi}_i) + (1-\lambda) \mathcal{L}(\vect{\pi}_j) $ for any $\lambda \in [0, 1]$.
This ensures that our scaler achieves Property \ref{prop2} and Property \ref{prop3}. The resulting global scaler, $\overline{\phi}(\vect{\pi})$, achieves a low local calibration error and is robust to the choice of aggregation weight $\lambda$. 
For practical implementation, we adopt the Weight Matching algorithm \cite{ainsworth2022git} which does not require access to local data (Algorithm \ref{alg:WeighMatching} in Appendix \ref{Ap:wm}). 

\subsection{The Combined Framework}

\begin{algorithm}[!t]
\caption{\methodname{}}
\label{alg:FedCal}
\begin{algorithmic}[1]
\STATE \textbf{Input:} $C$ clients, local epochs $E$, learning rate $\eta$, global model weights $\vect{\theta}^0$, global scaler weights $\overline{\vect{\pi}}^0$
\FOR{each round $t = 1, 2, \ldots, R$}
    \STATE Server selects a subset of $m$ clients $S_t$
    \FOR{each client $c \in S_t$ \textbf{in parallel}}
        \STATE $\vect{\theta}_c^{t+1}, \vect{\pi}_c^{t+1} \leftarrow \text{ClientUpdate}(c, \vect{\theta}^0, \overline{\vect{\pi}}^0)$
    \ENDFOR
    \STATE $\vect{\theta}^{t+1} \leftarrow \sum_{c=1}^{m} \frac{n_c}{N} \vect{\theta}_c^{t+1}$ \COMMENT{Aggregate updated models}
    \STATE \textcolor{blue}{$\overline{\vect{\pi}}^{t+1} \leftarrow \frac{1}{m}\sum_{c=1}^{m}\vect{\pi}_c^{t+1}$\COMMENT{Aggregate updated scalers}}
\ENDFOR 
\STATE \hrulefill
\STATE \textbf{procedure} 
\text{ClientUpdate}($(c, \vect{\theta}^0, \overline{\vect{\pi}}^0)$):
    \STATE 
    \textcolor{blue}{
    $\mathbf{M} \leftarrow \text{WeightMatching}(\overline{\vect{\pi}}^0,
    \vect{\pi}_c)$}
    \STATE 
    \textcolor{blue}{
    $\vect{\pi}_c\leftarrow \mathbf{M}(\vect{\pi}_c)$\COMMENT{Permute to Align}}
    \STATE $\mathcal{B} \leftarrow$ (split $\mathcal{D}_c$ into batches of size $B$)
    \FOR{each local epoch $i$ from 1 to $E$}
        \FOR{batch $b \in \mathcal{B}$}
            \STATE $w \leftarrow w - \eta \nabla l(w; b)$ \COMMENT{Update model weights}
        \ENDFOR
    \ENDFOR
    \STATE
    \textcolor{blue}{
     $\vect{\pi}_c \leftarrow \text{TrainScaler}(\vect{\pi}_c, \mathcal{D}_c)$}
    \STATE \textbf{return} $w$, $\vect{\pi}_c$ to server
\end{algorithmic}
\end{algorithm}

\methodname{} (Algorithm \ref{alg:FedCal}) enhances the FedAvg procedure to improve model calibration. Initially, FL clients are provided with the global scaler parameters as a baseline for local scaler training. To establish a coherent relationship between their local and the global scaler, clients perform Weight Matching (Algorithm \ref{alg:WeighMatching}) to achieve optimal alignment.
After local model updates, clients refine their scalers to minimize local calibration errors on their validation datasets. Upon receiving the updated models and scalers from the clients, the FL server performs model aggregation via FedAvg and averages the local scalers to form an updated global scaler.

Notably, rather than adopting a purely post-hoc approach, where scaler training and aggregation occur only after the completion of FedAvg, \methodname{} opts for periodic synchronization of scaler updates. This strategy enhances the scalers' ability to learn more general mappings and boost overall aggregation efficacy.
\methodname{} incurs additional communication overhead, equivalent to the scaler parameter count, and extra computational load due to weight matching. Nonetheless, given that the scaler parameter count is relatively modest compared to that of the primary classifier, the trade-off is considered acceptable.

Regarding privacy, \methodname{} necessitates the sharing of scaler parameters, which helps prevent the disclosure of specific data distribution details (such as logits, quantiles, etc.). Moreover, sharing parameters aligns with the standard paradigm in federated learning, making privacy-preserving techniques, such as Homomorphic Encryption \citep{zhang2020batchcrypt, hardy2017private} and Differential Privacy \citep{wei2020federated}, could be integrated to augment privacy protection further. However, the incorporation and exploration of these techniques fall outside the scope of our current research.


\section{Experimental Evaluation}
To evaluate the performance of \methodname{}, we experimentally compare it with five baseline methods over four benchmark datasets with different degrees of non-IIDness.

\subsection{Experiment Setup}
We conduct our experiments on the following widely adopted benchmark datasets: 1) MNIST~\citep{deng2012mnist}, 2) SVHN~\citep{37648}, 3) CIFAR-10 and 4) CIFAR-100~\cite{krizhevsky2009learning}. For MNIST and SVHN, we utilize the standard CNN as the base model, while for CIFAR-10 and CIFAR-100, we adopt ResNet-14 \citep{he2016deep} and ResNet-32 \citep{he2016deep} as the base model, respectively. Prior to distribution among FL clients, each dataset is pre-processed.

To replicate non-IID conditions typical in real-world settings, we follow the Distribution-based Label Skew method~\cite{yurochkin2019bayesian, li2020practical, zhang2022federated}, which uses a Dirichlet distribution ($p_c \sim \text{Dir}(\beta)$) to allocate class samples across clients. The parameter $\beta$ modulates label skew, with higher values indicating more pronounced non-IIDness. 

We carried out our experiments using PyTorch on a single NVIDIA A100 GPU, which has 40 GB of memory. In our FL setup, we include 20 FL clients. In each FL training round, we randomly select 5 of them to participate. Each local training round consists of 3 epochs. For the local updates, we adopt the SGD optimizer with a learning rate of 0.01, and a local batch size of 256. For the implementation of \methodname{}, the default configuration of the proposed scaler is an MLP with an activation structure of $K$-64-64-$K$, where $K$ represents the number of classes. We set the maximum number of global epochs to 100. 

\begin{table*}[ht]
\begin{center}

\label{tab:main_res}

\begin{small}
\begin{sc}
\resizebox*{1\linewidth}{!}{
\begin{tabular}{|l|ll|ll|lll|l|}
\hline

\multirow{2}{*}{Datasets}          & \multicolumn{2}{c|}{\textbf{Settings}}                        & \multicolumn{6}{|c|}{\textbf{Global ECE \%}}   

\\
\cline{2-9}

                          &   $\beta$  & Acc \%                     & Uncal. & Val. TS & Ens.  & AvgT.  & LR-TS.  & \methodname{}  \\
\hline
\multirow{4}{*}{MNIST-cnn}         & $\beta$ =1   & 95.72 {\tiny$\pm$ 2.3}   & 0.5 {\tiny$\pm$ 0.12}   & 0.34 {\tiny$\pm$ 0.08}   & 0.43 {\tiny$\pm$ 0.10} & \textbf{0.41} {\tiny$\pm$ 0.10}  & 0.71 {\tiny$\pm$ 0.17}   & 0.45 {\tiny$\pm$ 0.11}   \\
                                   & $\beta$ =0.5 & 93.03 {\tiny$\pm$ 2.2}   & 1.00 {\tiny$\pm$ 0.24}  & 0.79 {\tiny$\pm$ 0.19}   & 1.32 {\tiny$\pm$ 0.32} & 0.72 {\tiny$\pm$ 0.17} & 0.71 {\tiny$\pm$ 0.17}   & \textbf{0.47} {\tiny$\pm$ 0.11}   \\
                                   & $\beta$ =0.3 & 91.34 {\tiny$\pm$ 2.2}   & 1.6 {\tiny$\pm$ 0.38}   & 0.74 {\tiny$\pm$ 0.18}   & 1.34 {\tiny$\pm$ 0.32} & 0.92 {\tiny$\pm$ 0.22} & 1.57 {\tiny$\pm$ 0.38}   & \textbf{0.61} {\tiny$\pm$ 0.15}  \\
                                   & $\beta$ =0.1 & 81.01 {\tiny$\pm$ 1.9}   & 4.6 {\tiny$\pm$ 1.10}   & 0.74 {\tiny$\pm$ 0.18}   & 2.13 {\tiny$\pm$ 0.51} & 3.17 {\tiny$\pm$ 0.76} & 5.51 {\tiny$\pm$ 1.32}   & \textbf{1.35} {\tiny$\pm$ 0.32}  \\
\hline

\multirow{4}{*}{SVNH-cnn}          & $\beta$=1   & 93.24 {\tiny$\pm$ 2.2}   & 1.12 {\tiny$\pm$ 0.27}  & 0.32 {\tiny$\pm$ 0.08}   & 0.43 {\tiny$\pm$ 0.10} & 0.52 {\tiny$\pm$ 0.12} & 0.69 {\tiny$\pm$ 0.17}   & \textbf{0.44} {\tiny$\pm$ 0.11}   \\
                                   & $\beta$=0.5 & 85.13 {\tiny$\pm$ 2.0}   & 1.30 {\tiny$\pm$ 0.31}  & 0.51 {\tiny$\pm$ 0.12}   & 0.99 {\tiny$\pm$ 0.24} & 1.03 {\tiny$\pm$ 0.25} & 1.31 {\tiny$\pm$ 0.31}   & \textbf{0.89} {\tiny$\pm$ 0.21}   \\
                                   & $\beta$=0.3 & 85.14 {\tiny$\pm$ 2.0}   & 4.56 {\tiny$\pm$ 1.09}  & 1.21 {\tiny$\pm$ 0.29}   & 1.58 {\tiny$\pm$ 0.38} & 3.59 {\tiny$\pm$ 0.86} & 3.51 {\tiny$\pm$ 0.84}   & \textbf{0.77} {\tiny$\pm$ 0.18}  \\
                                   & $\beta$=0.1 & 79.23 {\tiny$\pm$ 1.9}   & 7.81 {\tiny$\pm$ 1.87}  & 1.25 {\tiny$\pm$ 0.30}   & 1.56 {\tiny$\pm$ 0.37} & 6.12 {\tiny$\pm$ 1.47} & 32.03 {\tiny$\pm$ 7.69}  & \textbf{1.25} {\tiny$\pm$ 0.30}   \\
\hline

\multirow{4}{*}{CIFAR10-resnet14}  & $\beta$=1   & 65.54 {\tiny$\pm$ 1.6}   & 7.61 {\tiny$\pm$ 1.82}  & 3.61 {\tiny$\pm$ 0.87}   & 5.42 {\tiny$\pm$ 1.30} & 7.82 {\tiny$\pm$ 1.88} & 3.42 {\tiny$\pm$ 0.82}   & \textbf{3.71} {\tiny$\pm$ 0.89}  \\
                                   & $\beta$=0.5 & 60.21 {\tiny$\pm$ 1.4}   & 5.63 {\tiny$\pm$ 1.35}  & 4.12 {\tiny$\pm$ 0.99}   & 6.28 {\tiny$\pm$ 1.51} & 8.34 {\tiny$\pm$ 2.00} & 4.23 {\tiny$\pm$ 1.01}   & \textbf{4.61} {\tiny$\pm$ 1.10}   \\
                                   & $\beta$=0.3 & 57.31 {\tiny$\pm$ 1.4}   & 9.81 {\tiny$\pm$ 2.35}  & 3.15 {\tiny$\pm$ 0.76}   & 8.16 {\tiny$\pm$ 1.96} & 11.25 {\tiny$\pm$ 2.70} & 11.12 {\tiny$\pm$ 2.67}  & \textbf{4.91} {\tiny$\pm$ 1.18}   \\
                                   & $\beta$=0.1 & 48.05 {\tiny$\pm$ 1.2}   & 12.48 {\tiny$\pm$ 2.99} & 3.51 {\tiny$\pm$ 0.84}   & 8.87 {\tiny$\pm$ 2.13} & 13.34 {\tiny$\pm$ 3.20} & 14.32 {\tiny$\pm$ 3.44} & \textbf{4.51} {\tiny$\pm$ 1.08}   \\
\hline

\multirow{4}{*}{CIFAR100-resnet32} & $\beta$=1   & 41.24 {\tiny$\pm$ 1.0}   & 22.45 {\tiny$\pm$ 5.39} & 4.13 {\tiny$\pm$ 0.99}   & 16.34 {\tiny$\pm$ 3.92} & 20.25 {\tiny$\pm$ 4.86} & 19.21 {\tiny$\pm$ 4.61}  & \textbf{7.41} {\tiny$\pm$ 1.78}  \\
                                   & $\beta$=0.5 & 30.01 {\tiny$\pm$ 0.7}   & 20.45 {\tiny$\pm$ 4.91} & 4.19 {\tiny$\pm$ 1.01}   & 17.53 {\tiny$\pm$ 4.21} & 29.21 {\tiny$\pm$ 7.01} & 18.35 {\tiny$\pm$ 4.40}  & \textbf{7.50} {\tiny$\pm$ 1.80}   \\
                                   & $\beta$=0.3 & 22.21 {\tiny$\pm$ 0.5}   & 25.71 {\tiny$\pm$ 6.17} & 4.09 {\tiny$\pm$ 0.98}   & 18.93 {\tiny$\pm$ 4.55} & 30.45 {\tiny$\pm$ 7.31} & 22.92 {\tiny$\pm$ 5.50}  & \textbf{8.91} {\tiny$\pm$ 2.14}   \\
                                   & $\beta$=0.1 & 20.8 {\tiny$\pm$ 0.5}    & 31.78 {\tiny$\pm$ 7.63} & 4.82 {\tiny$\pm$ 1.16}   & 20.48 {\tiny$\pm$ 4.92} & 30.25 {\tiny$\pm$ 7.26} & 37.32 {\tiny$\pm$ 8.96} & \textbf{10.72} {\tiny$\pm$ 2.57}  \\

\hline
\end{tabular}
}
\end{sc}
\end{small}
\end{center}
\caption{Comparison of global ECE across datasets with varying levels of non-IIDness.}
\end{table*}

\subsection{Comparison Baselines}

We compare \methodname{} with standard scaler designs and FL aggregation methods, including:
\begin{enumerate}
    \item \textsc{Uncal.}: FL without model calibration.
    \item \textsc{Val. TS}: uses a temperature scaler on a global validation set, considered as the performance upper bound.
    \item \textsc{Ens.}: Implements Deep Ensemble's direct averaging of scaled probabilities. Despite its simplicity, it's known for robust uncertainty quantification \citep{lakshminarayanan2017simple, lee2015m}.
    \item \textsc{AvgT}: Extends \textsc{Ens} by averaging temperature parameters of individual scalers. Both \textsc{Ens} and \textsc{AvgT} are efficient, but have limitations when facing non-IID distributions \cite{rahaman2021uncertainty, abe2022deep}.
    \item \textsc{LR-TS}: Adapts MD-TS \cite{yu2022robust} for FL settings. It estimates the scaling temperature from model outputs, thereby achieving calibration without a shared validation dataset.
\end{enumerate}
We adopt test accuracy and the global ECE as the evaluation metrics.
For more details, please refer to Appendix \ref{Ap:basline}.

\subsection{Results and Discussion}
Table \ref{tab:main_res} reports the global ECE results. 
\methodname{} consistently achieves the lowest calibration error across all datasets. When compared to \textsc{Uncal} and the second-best approach without requiring a global validation set \textsc{Ens.}, \methodname{} significantly reduces the calibration error by 63.06\% and 47.66\%, respectively. It is also important to note that an increase in non-IIDness tends to worsen the calibration error, which corroborates our theoretical analysis. However, \methodname{} demonstrates stronger robustness compared to other baselines, even as non-IIDness increases. In most cases, \textsc{Ens} emerges as the second-best, reinforcing the notion that deep ensembles are effective in calibration, particularly when non-IIDness is moderate.

\begin{figure}[!h]
\begin{center}
\centerline{\includegraphics[width=1\linewidth]{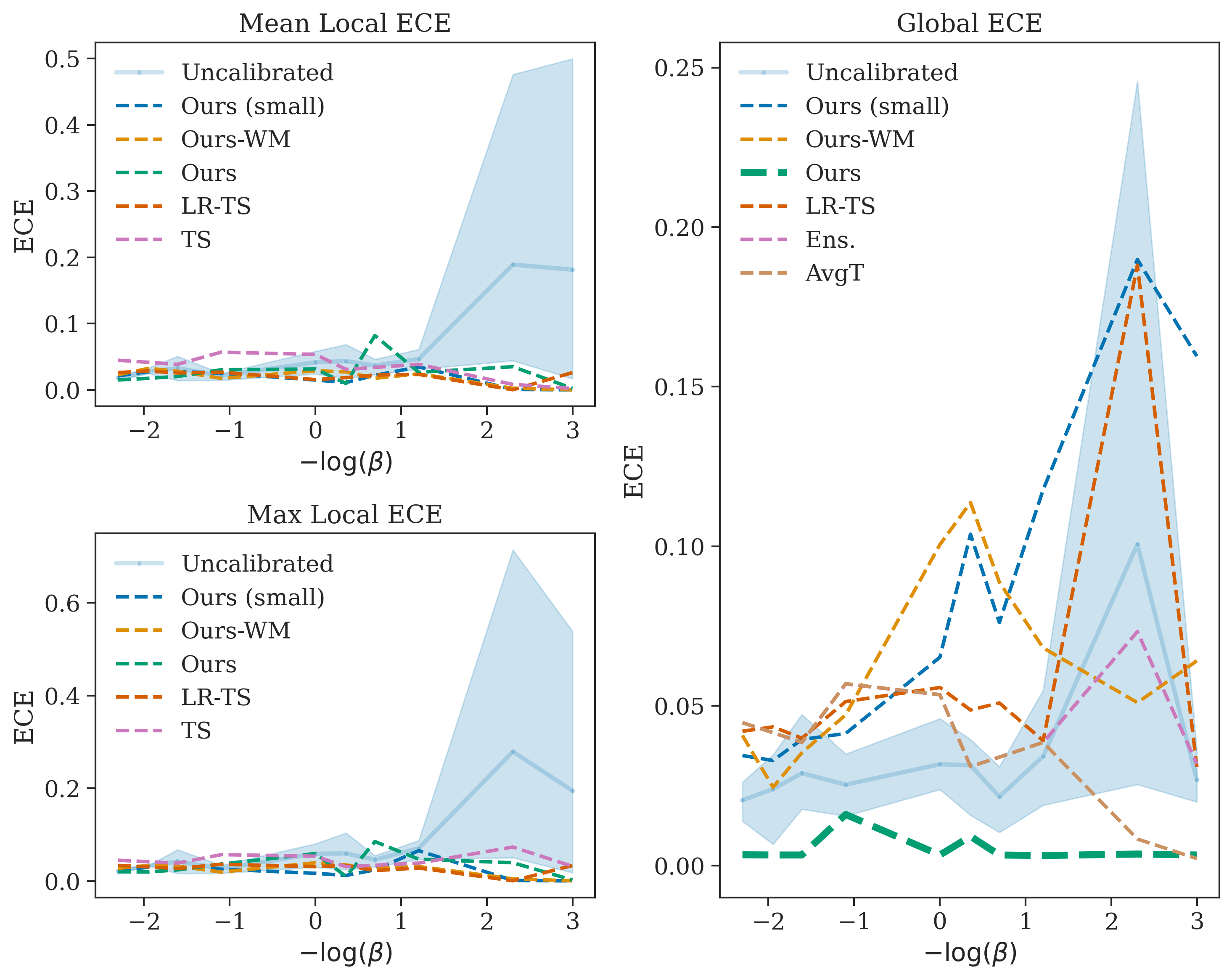}}
\caption{Local and global calibration errors as non-IIDness increases. [Top left]: the average local calibration errors. [Bottom left]: the maximum local calibration errors. [Right]: the global calibration error.}
\label{fig: ablation-studies}
\vspace{-10pt}
\end{center}
\end{figure}

Under MNIST, we examine how calibration error changes with increasing non-IIDness (as indicated by $-\log\beta$). The results are illustrated in Figure \ref{fig: ablation-studies}, which also contains the results of parts of our ablation studies. 
The top left subplot indicates the average local calibration error, while the bottom left one shows the maximum local calibration error. It can be observed that despite the increase in non-IIDness, all scaling methods manage to maintain low calibration errors.

In contrast, the global calibration error Figure \ref{fig: ablation-studies} (right sub-plot) behaves differently. It can be observed that \methodname{} consistently outperforms other methods without a substantial increase in global ECE as non-IIDness grows. However, without performing weight matching (\textsc{Ours-WM}), \methodname{} does not exhibit the same robustness. Similarly, reducing the MLP scaler size from 64 neurons to 8 (\textsc{Ours-small}) results in a reduction in the generalization capabilities of \methodname{}. This underscores the importance of the synergy of the integrated scaler and the aggregation strategy in the \methodname{} design.

\subsection{Ablation Studies}

\begin{figure}[!t]
\begin{center}
\centerline{\includegraphics[width=1\linewidth]{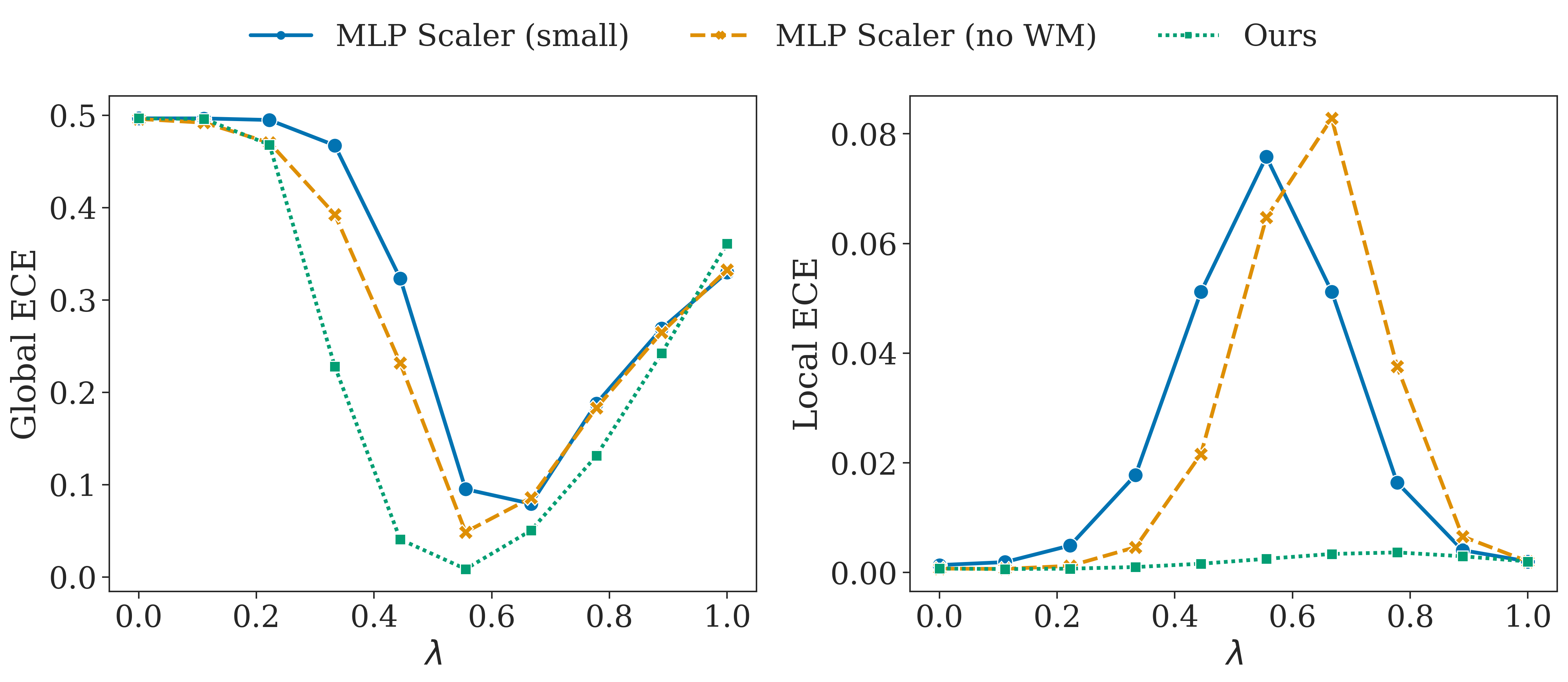}}
\caption{
Local and global ECE vs. aggregation weights.} 
\label{fig:weightmatching}
\vspace{-10 pt}
\end{center}
\end{figure}

In our ablation studies, we demonstrate the effectiveness of MLP scalers and weight matching in \methodname{} (Figure \ref{fig:weightmatching}). Using a synthetic dataset with high non-IIDness and two clients, we find that while individual local scalers are inadequate globally, their weighted aggregation significantly improves overall calibration. Increasing the MLP scaler size also enhances generalization. Directly applying global scalers locally leads to good individual calibration, but causes substantial local errors without a proper weight matching mechanism. The results highlight the importance of weight matching in federated calibration.

In \methodname{}, the order-preserving technique is a crucial component. We evaluate its significance through ablation studies conducted on the MNIST dataset, where $\beta$ is fixed at 0.5. The results are shown in Table \ref{tab:top-3}. 

It can be observed that calibration methods like \textsc{Ens}, \textsc{AvgT} and \textsc{LR-TS}, which modify model logits, have no detrimental effect on top-3 accuracy. In contrast, removing the order-preserving element from \methodname{} (denoted as \textsc{Ours w/o OP}) results in a slight reduction in global ECE. Yet, this comes at the expense of significantly lower top-3 accuracy. This outcome suggests that, while \methodname{} can enhance test accuracy to a certain extent, excluding the order-preserving part of the design negatively affects the model's ability to accurately rank its top predictions. Conversely, while imposing order-preserving constraints does introduce some limitations, it only leads to a minor increase in global ECE.

\begin{table}[ht]
\label{tab:top-3}
\begin{center}
\begin{small}
\begin{sc}
\begin{tabular}{lcccr}
\toprule
               & Global ECE & Top-3 Accuracy\\\midrule
Without Scaler & 4.6\%               & \underline{92.3\%}              \\
Ens.         & 2.13\%                 & -                 \\
AvgT             & 3.17\%                & -                 \\
LR-TS          & 5.51\%                & -                 \\
\midrule
Ours w/o OP         & \textbf{1.32\%}        & 87.91\%             \\
Ours            & \underline{1.71\%}         & \textbf{92.3\%}    \\
\bottomrule
\end{tabular}
\end{sc}
\end{small}
\end{center}
\caption{Top-3 Accuracy and Calibration Error on the MNIST dataset with a CNN with $\beta = 0.5$.}
\end{table}

We also evaluated \methodname{} in combination with methods designed to improve federated learning performance under non-IID data distributions. Specifically, we tested\methodname{} together with FedProx \citep{li2020federated}, which adds regularization to encourage local updates to stay closer to the global model under non-IID conditions.
As shown in Table \ref{tab:FedProx}, while FedProx improves accuracy, it still suffers from significant calibration errors. However, combining it with \methodname{} substantially reduced global calibration errors. This demonstrates that \methodname{} effectively complements existing non-IID approaches (e.g., FedProx) by enhancing calibration.

\begin{table}[h]

\label{tab:FedProx}
\begin{center}
\begin{small}
\begin{sc}
\begin{tabular}{c c || c | c ||}
\hline
\toprule
 $\multirow{2}{*}{$\beta$}$ & \multirow{2}{*}{Acc \%}& \multicolumn{2}{c||}{\textbf{Global ECE \%}}   \\ 
 & & FedProx & FedProx + \methodname{} \\
\midrule
$\beta$ =1   & 94.32 & 0.38 & 0.40 \\ 
$\beta$ = 0.5 & 94.07 & 0.57 & 0.42 \\ 
$\beta$ = 0.3 & 90.25 & 1.32 & 0.58 \\ 
$\beta$ = 0.1 & 89.93 & 3.19 & 1.07 \\ \bottomrule
\end{tabular}
\end{sc}
\end{small}
\end{center}
\caption{Global calibration errors on the MNIST-CNN for FedAvg, FedProx, and FedProx + FedCal under varying non-IID levels}
\end{table}

\section{Conclusions and Future Work}
In this paper, we provide both theoretical and empirical insights into the necessity of simultaneously achieving global calibration and local calibration in FL settings. The proposed \methodname{} approach is designed with a sophisticated multi-layer perceptron (MLP) scaler alongside the order-preserving technique to effectively handle the challenges posed by non-IID data distributions commonly encountered in real-world FL applications.
Extensive experiments demonstrate that \methodname{} surpasses existing calibration methods by significantly reducing global model calibration error without compromising model accuracy.

In subsequent research, we plan to investigate the complex dynamics between order-preserving techniques and weight matching strategies. We will also develop theories about the relations between calibration effectiveness and model size. 

\section*{Acknowledgements}
This research is supported, in part, by the National Research Foundation Singapore and DSO National Laboratories under the AI Singapore Programme (No. AISG2-RP-2020-019); the RIE2025 Industry Alignment Fund – Industry Collaboration Projects (IAF-ICP) (Award I2301E0026), administered by A*STAR, as well as supported by Alibaba Group and NTU Singapore; the Natural Sciences and Engineering Research Council of Canada (NSERC); and Compute Canada Research Platform.


\section*{Impact Statement}
This paper presents work whose goal is to advance the field of Machine Learning. There are many potential societal consequences of our work, none of which we feel must be specifically highlighted here.

\bibliography{ref}

\begin{thebibliography}{63}
\providecommand{\natexlab}[1]{#1}
\providecommand{\url}[1]{\texttt{#1}}
\expandafter\ifx\csname urlstyle\endcsname\relax
  \providecommand{\doi}[1]{doi: #1}\else
  \providecommand{\doi}{doi: \begingroup \urlstyle{rm}\Url}\fi

\bibitem[Abe et~al.(2022)Abe, Buchanan, Pleiss, Zemel, and Cunningham]{abe2022deep}
Abe, T., Buchanan, E.~K., Pleiss, G., Zemel, R., and Cunningham, J.~P.
\newblock Deep ensembles work, but are they necessary?
\newblock \emph{Advances in Neural Information Processing Systems}, 35:\penalty0 33646--33660, 2022.

\bibitem[Achituve et~al.(2021)Achituve, Shamsian, Navon, Chechik, and Fetaya]{fl-calibration-achituve2021personalized}
Achituve, I., Shamsian, A., Navon, A., Chechik, G., and Fetaya, E.
\newblock Personalized federated learning with gaussian processes.
\newblock \emph{Advances in Neural Information Processing Systems}, 34:\penalty0 8392--8406, 2021.

\bibitem[Ainsworth et~al.(2022)Ainsworth, Hayase, and Srinivasa]{ainsworth2022git}
Ainsworth, S.~K., Hayase, J., and Srinivasa, S.
\newblock Git re-basin: Merging models modulo permutation symmetries.
\newblock \emph{arXiv preprint arXiv:2209.04836}, 2022.

\bibitem[Angelopoulos et~al.(2021)Angelopoulos, Bates, Cand{\`e}s, Jordan, and Lei]{angelopoulos2021learn}
Angelopoulos, A.~N., Bates, S., Cand{\`e}s, E.~J., Jordan, M.~I., and Lei, L.
\newblock Learn then test: Calibrating predictive algorithms to achieve risk control.
\newblock \emph{arXiv preprint arXiv:2110.01052}, 2021.

\bibitem[Arivazhagan et~al.(2019)Arivazhagan, Aggarwal, Singh, and Choudhary]{arivazhagan2019federated}
Arivazhagan, M.~G., Aggarwal, V., Singh, A.~K., and Choudhary, S.
\newblock Federated learning with personalization layers.
\newblock \emph{arXiv preprint arXiv:1912.00818}, 2019.

\bibitem[Charles \& Kone{\v{c}}n{\`y}(2021)Charles and Kone{\v{c}}n{\`y}]{charles2021convergence}
Charles, Z. and Kone{\v{c}}n{\`y}, J.
\newblock Convergence and accuracy trade-offs in federated learning and meta-learning.
\newblock In \emph{International Conference on Artificial Intelligence and Statistics}, pp.\  2575--2583. PMLR, 2021.

\bibitem[Dash et~al.(2022)Dash, Sharma, and Ali]{dash2022federated}
Dash, B., Sharma, P., and Ali, A.
\newblock Federated learning for privacy-preserving: A review of pii data analysis in fintech.
\newblock \emph{International Journal of Software Engineering \& Applications (IJSEA)}, 13\penalty0 (4), 2022.

\bibitem[Dayan et~al.(2021)Dayan, Roth, Zhong, Harouni, Gentili, Abidin, Liu, Costa, Wood, Tsai, et~al.]{dayan2021federated}
Dayan, I., Roth, H.~R., Zhong, A., Harouni, A., Gentili, A., Abidin, A.~Z., Liu, A., Costa, A.~B., Wood, B.~J., Tsai, C.-S., et~al.
\newblock Federated learning for predicting clinical outcomes in patients with covid-19.
\newblock \emph{Nature medicine}, 27\penalty0 (10):\penalty0 1735--1743, 2021.

\bibitem[Deng(2012)]{deng2012mnist}
Deng, L.
\newblock The mnist database of handwritten digit images for machine learning research.
\newblock \emph{IEEE Signal Processing Magazine}, 29\penalty0 (6):\penalty0 141--142, 2012.

\bibitem[Deng et~al.(2020)Deng, Kamani, and Mahdavi]{deng2020adaptive}
Deng, Y., Kamani, M.~M., and Mahdavi, M.
\newblock Adaptive personalized federated learning.
\newblock \emph{arXiv preprint arXiv:2003.13461}, 2020.

\bibitem[Du et~al.(2020)Du, Wu, Yoshinaga, Yau, Ji, and Li]{du2020federated}
Du, Z., Wu, C., Yoshinaga, T., Yau, K.-L.~A., Ji, Y., and Li, J.
\newblock Federated learning for vehicular internet of things: Recent advances and open issues.
\newblock \emph{IEEE Open Journal of the Computer Society}, 1:\penalty0 45--61, 2020.

\bibitem[Entezari et~al.(2021)Entezari, Sedghi, Saukh, and Neyshabur]{entezari2021role}
Entezari, R., Sedghi, H., Saukh, O., and Neyshabur, B.
\newblock The role of permutation invariance in linear mode connectivity of neural networks.
\newblock \emph{arXiv preprint arXiv:2110.06296}, 2021.

\bibitem[Guo et~al.(2017{\natexlab{a}})Guo, Pleiss, Sun, and Weinberger]{Guo_Pleiss_Sun_Weinberger_2017}
Guo, C., Pleiss, G., Sun, Y., and Weinberger, K.~Q.
\newblock On calibration of modern neural networks.
\newblock In \emph{International conference on machine learning}, pp.\  1321--1330. PMLR, 2017{\natexlab{a}}.

\bibitem[Guo et~al.(2017{\natexlab{b}})Guo, Pleiss, Sun, and Weinberger]{guo2017calibration}
Guo, C., Pleiss, G., Sun, Y., and Weinberger, K.~Q.
\newblock On calibration of modern neural networks.
\newblock In \emph{International conference on machine learning}, pp.\  1321--1330. PMLR, 2017{\natexlab{b}}.

\bibitem[Hardy et~al.(2017)Hardy, Henecka, Ivey-Law, Nock, Patrini, Smith, and Thorne]{hardy2017private}
Hardy, S., Henecka, W., Ivey-Law, H., Nock, R., Patrini, G., Smith, G., and Thorne, B.
\newblock Private federated learning on vertically partitioned data via entity resolution and additively homomorphic encryption.
\newblock \emph{arXiv preprint arXiv:1711.10677}, 2017.

\bibitem[He et~al.(2016)He, Zhang, Ren, and Sun]{he2016deep}
He, K., Zhang, X., Ren, S., and Sun, J.
\newblock Deep residual learning for image recognition.
\newblock In \emph{Proceedings of the IEEE conference on computer vision and pattern recognition}, pp.\  770--778, 2016.

\bibitem[Hendrycks et~al.(2019)Hendrycks, Mazeika, Kadavath, and Song]{hendrycks2019using}
Hendrycks, D., Mazeika, M., Kadavath, S., and Song, D.
\newblock Using self-supervised learning can improve model robustness and uncertainty.
\newblock \emph{Advances in neural information processing systems}, 32, 2019.

\bibitem[Hsieh et~al.(2020)Hsieh, Phanishayee, Mutlu, and Gibbons]{hsieh2020non}
Hsieh, K., Phanishayee, A., Mutlu, O., and Gibbons, P.
\newblock The non-iid data quagmire of decentralized machine learning.
\newblock In \emph{International Conference on Machine Learning}, pp.\  4387--4398. PMLR, 2020.

\bibitem[Kairouz et~al.(2021)Kairouz, McMahan, Avent, Bellet, Bennis, Bhagoji, Bonawitz, Charles, Cormode, Cummings, et~al.]{kairouz2021advances}
Kairouz, P., McMahan, H.~B., Avent, B., Bellet, A., Bennis, M., Bhagoji, A.~N., Bonawitz, K., Charles, Z., Cormode, G., Cummings, R., et~al.
\newblock Advances and open problems in federated learning.
\newblock \emph{Foundations and Trends{\textregistered} in Machine Learning}, 14\penalty0 (1--2):\penalty0 1--210, 2021.

\bibitem[Karimireddy et~al.(2020)Karimireddy, Kale, Mohri, Reddi, Stich, and Suresh]{karimireddy2020scaffold}
Karimireddy, S.~P., Kale, S., Mohri, M., Reddi, S., Stich, S., and Suresh, A.~T.
\newblock Scaffold: Stochastic controlled averaging for federated learning.
\newblock In \emph{International conference on machine learning}, pp.\  5132--5143. PMLR, 2020.

\bibitem[Kong et~al.(2020)Kong, Sun, and Zhang]{kong2020sde}
Kong, L., Sun, J., and Zhang, C.
\newblock Sde-net: Equipping deep neural networks with uncertainty estimates.
\newblock \emph{arXiv preprint arXiv:2008.10546}, 2020.

\bibitem[Krizhevsky et~al.(2009)Krizhevsky, Hinton, et~al.]{krizhevsky2009learning}
Krizhevsky, A., Hinton, G., et~al.
\newblock Learning multiple layers of features from tiny images.
\newblock Technical report, University of Toronto, Canada, 2009.

\bibitem[Kumar \& Sarawagi(2019)Kumar and Sarawagi]{kumar2019calibration}
Kumar, A. and Sarawagi, S.
\newblock Calibration of encoder decoder models for neural machine translation.
\newblock \emph{arXiv preprint arXiv:1903.00802}, 2019.

\bibitem[Kumar et~al.(2019)Kumar, Liang, and Ma]{kumar2019verified}
Kumar, A., Liang, P.~S., and Ma, T.
\newblock Verified uncertainty calibration.
\newblock \emph{Advances in Neural Information Processing Systems}, 32, 2019.

\bibitem[Lakshminarayanan et~al.(2017)Lakshminarayanan, Pritzel, and Blundell]{lakshminarayanan2017simple}
Lakshminarayanan, B., Pritzel, A., and Blundell, C.
\newblock Simple and scalable predictive uncertainty estimation using deep ensembles.
\newblock \emph{Advances in neural information processing systems}, 30, 2017.

\bibitem[Lee et~al.(2015)Lee, Purushwalkam, Cogswell, Crandall, and Batra]{lee2015m}
Lee, S., Purushwalkam, S., Cogswell, M., Crandall, D., and Batra, D.
\newblock Why m heads are better than one: Training a diverse ensemble of deep networks.
\newblock \emph{arXiv preprint arXiv:1511.06314}, 2015.

\bibitem[Li et~al.(2020{\natexlab{a}})Li, Wen, and He]{li2020practical}
Li, Q., Wen, Z., and He, B.
\newblock Practical federated gradient boosting decision trees.
\newblock In \emph{Proceedings of the AAAI conference on artificial intelligence}, volume~34, pp.\  4642--4649, 2020{\natexlab{a}}.

\bibitem[Li et~al.(2020{\natexlab{b}})Li, Sahu, Zaheer, Sanjabi, Talwalkar, and Smith]{li2020federated}
Li, T., Sahu, A.~K., Zaheer, M., Sanjabi, M., Talwalkar, A., and Smith, V.
\newblock Federated optimization in heterogeneous networks.
\newblock \emph{Proceedings of Machine learning and systems}, 2:\penalty0 429--450, 2020{\natexlab{b}}.

\bibitem[Li et~al.(2021)Li, Tao, Zhang, Liu, and Xu]{li2021privacy}
Li, Y., Tao, X., Zhang, X., Liu, J., and Xu, J.
\newblock Privacy-preserved federated learning for autonomous driving.
\newblock \emph{IEEE Transactions on Intelligent Transportation Systems}, 23\penalty0 (7):\penalty0 8423--8434, 2021.

\bibitem[Liu et~al.(2024)Liu, Xing, Deng, Li, Guan, and Yu]{liu2024federated}
Liu, R., Xing, P., Deng, Z., Li, A., Guan, C., and Yu, H.
\newblock Federated graph neural networks: Overview, techniques, and challenges.
\newblock \emph{IEEE Transactions on Neural Networks and Learning Systems}, 2024.

\bibitem[Long et~al.(2020)Long, Tan, Jiang, and Zhang]{long2020federated}
Long, G., Tan, Y., Jiang, J., and Zhang, C.
\newblock Federated learning for open banking.
\newblock In \emph{Federated Learning: Privacy and Incentive}, pp.\  240--254. Springer, 2020.

\bibitem[Lu \& Kalpathy-Cramer(2021)Lu and Kalpathy-Cramer]{lu2021distribution}
Lu, C. and Kalpathy-Cramer, J.
\newblock Distribution-free federated learning with conformal predictions.
\newblock \emph{arXiv preprint arXiv:2110.07661}, 2021.

\bibitem[Luo et~al.(2019)Luo, Wu, Luo, Huang, Huang, Liu, and Yang]{luo2019real}
Luo, J., Wu, X., Luo, Y., Huang, A., Huang, Y., Liu, Y., and Yang, Q.
\newblock Real-world image datasets for federated learning.
\newblock \emph{arXiv preprint arXiv:1910.11089}, 2019.

\bibitem[Luo et~al.(2021)Luo, Chen, Hu, Zhang, Liang, and Feng]{fl-calibration-luo2021no}
Luo, M., Chen, F., Hu, D., Zhang, Y., Liang, J., and Feng, J.
\newblock No fear of heterogeneity: Classifier calibration for federated learning with non-iid data.
\newblock \emph{Advances in Neural Information Processing Systems}, 34:\penalty0 5972--5984, 2021.

\bibitem[Lyu et~al.(2022)Lyu, Yu, Ma, Chen, Sun, Zhao, Yang, and Philip]{lyu2022privacy}
Lyu, L., Yu, H., Ma, X., Chen, C., Sun, L., Zhao, J., Yang, Q., and Philip, S.~Y.
\newblock Privacy and robustness in federated learning: Attacks and defenses.
\newblock \emph{IEEE transactions on neural networks and learning systems}, 2022.

\bibitem[McMahan et~al.(2017)McMahan, Moore, Ramage, Hampson, and y~Arcas]{McMahan_Moore_Ramage_Hampson_Arcas_2017}
McMahan, B., Moore, E., Ramage, D., Hampson, S., and y~Arcas, B.~A.
\newblock Communication-efficient learning of deep networks from decentralized data.
\newblock In \emph{Artificial intelligence and statistics}, pp.\  1273--1282. PMLR, 2017.

\bibitem[Naeini et~al.(2015)Naeini, Cooper, and Hauskrecht]{naeini2015obtaining}
Naeini, M.~P., Cooper, G., and Hauskrecht, M.
\newblock Obtaining well calibrated probabilities using bayesian binning.
\newblock In \emph{Proceedings of the AAAI conference on artificial intelligence}, volume~29, 2015.

\bibitem[Netzer et~al.(2011)Netzer, Wang, Coates, Bissacco, Wu, and Ng]{37648}
Netzer, Y., Wang, T., Coates, A., Bissacco, A., Wu, B., and Ng, A.~Y.
\newblock Reading digits in natural images with unsupervised feature learning.
\newblock In \emph{NIPS Workshop on Deep Learning and Unsupervised Feature Learning 2011}, 2011.
\newblock URL \url{http://ufldl.stanford.edu/housenumbers/nips2011_housenumbers.pdf}.

\bibitem[Nguyen et~al.(2021)Nguyen, Br{\'e}chet, and Mondelli]{nguyen2021solutions}
Nguyen, Q.~N., Br{\'e}chet, P., and Mondelli, M.
\newblock When are solutions connected in deep networks?
\newblock \emph{Advances in Neural Information Processing Systems}, 34:\penalty0 20956--20969, 2021.

\bibitem[Ovadia et~al.(2019)Ovadia, Fertig, Ren, Nado, Sculley, Nowozin, Dillon, Lakshminarayanan, and Snoek]{Ovadia_Fertig_Ren_Nado_Sculley_Nowozin_Dillon_Lakshminarayanan_Snoek_2019}
Ovadia, Y., Fertig, E., Ren, J., Nado, Z., Sculley, D., Nowozin, S., Dillon, J., Lakshminarayanan, B., and Snoek, J.
\newblock Can you trust your model's uncertainty? evaluating predictive uncertainty under dataset shift.
\newblock \emph{Advances in neural information processing systems}, 32, 2019.

\bibitem[Plassier et~al.(2023)Plassier, Makni, Rubashevskii, Moulines, and Panov]{pmlr-v202-plassier23a}
Plassier, V., Makni, M., Rubashevskii, A., Moulines, E., and Panov, M.
\newblock Conformal prediction for federated uncertainty quantification under label shift.
\newblock In Krause, A., Brunskill, E., Cho, K., Engelhardt, B., Sabato, S., and Scarlett, J. (eds.), \emph{Proceedings of the 40th International Conference on Machine Learning}, volume 202 of \emph{Proceedings of Machine Learning Research}, pp.\  27907--27947. PMLR, 23--29 Jul 2023.
\newblock URL \url{https://proceedings.mlr.press/v202/plassier23a.html}.

\bibitem[Platt et~al.(1999)]{platt1999probabilistic}
Platt, J. et~al.
\newblock Probabilistic outputs for support vector machines and comparisons to regularized likelihood methods.
\newblock \emph{Advances in large margin classifiers}, 10\penalty0 (3):\penalty0 61--74, 1999.

\bibitem[Rahaman et~al.(2021)]{rahaman2021uncertainty}
Rahaman, R. et~al.
\newblock Uncertainty quantification and deep ensembles.
\newblock \emph{Advances in Neural Information Processing Systems}, 34:\penalty0 20063--20075, 2021.

\bibitem[Rahimi et~al.(2020)Rahimi, Shaban, Cheng, Hartley, and Boots]{rahimi2020intra}
Rahimi, A., Shaban, A., Cheng, C.-A., Hartley, R., and Boots, B.
\newblock Intra order-preserving functions for calibration of multi-class neural networks.
\newblock \emph{Advances in Neural Information Processing Systems}, 33:\penalty0 13456--13467, 2020.

\bibitem[Reddi et~al.(2020)Reddi, Charles, Zaheer, Garrett, Rush, Kone{\v{c}}n{\`y}, Kumar, and McMahan]{reddi2020adaptive}
Reddi, S., Charles, Z., Zaheer, M., Garrett, Z., Rush, K., Kone{\v{c}}n{\`y}, J., Kumar, S., and McMahan, H.~B.
\newblock Adaptive federated optimization.
\newblock \emph{arXiv preprint arXiv:2003.00295}, 2020.

\bibitem[Rieke et~al.(2020)Rieke, Hancox, Li, Milletari, Roth, Albarqouni, Bakas, Galtier, Landman, Maier-Hein, et~al.]{rieke2020future}
Rieke, N., Hancox, J., Li, W., Milletari, F., Roth, H.~R., Albarqouni, S., Bakas, S., Galtier, M.~N., Landman, B.~A., Maier-Hein, K., et~al.
\newblock The future of digital health with federated learning.
\newblock \emph{NPJ digital medicine}, 3\penalty0 (1):\penalty0 119, 2020.

\bibitem[Sheller et~al.(2020)Sheller, Edwards, Reina, Martin, Pati, Kotrotsou, Milchenko, Xu, Marcus, Colen, et~al.]{sheller2020federated}
Sheller, M.~J., Edwards, B., Reina, G.~A., Martin, J., Pati, S., Kotrotsou, A., Milchenko, M., Xu, W., Marcus, D., Colen, R.~R., et~al.
\newblock Federated learning in medicine: facilitating multi-institutional collaborations without sharing patient data.
\newblock \emph{Scientific reports}, 10\penalty0 (1):\penalty0 12598, 2020.

\bibitem[Tan et~al.(2023)Tan, Yu, Cui, and Yang]{tan2023towards}
Tan, A.~Z., Yu, H., Cui, L., and Yang, Q.
\newblock Towards personalized federated learning.
\newblock \emph{IEEE Transactions on Neural Networks and Learning Systems}, 34\penalty0 (12):\penalty0 9587--9603, 2023.

\bibitem[Thulasidasan et~al.(2019)Thulasidasan, Chennupati, Bilmes, Bhattacharya, and Michalak]{thulasidasan2019mixup}
Thulasidasan, S., Chennupati, G., Bilmes, J.~A., Bhattacharya, T., and Michalak, S.
\newblock On mixup training: Improved calibration and predictive uncertainty for deep neural networks.
\newblock \emph{Advances in Neural Information Processing Systems}, 32, 2019.

\bibitem[Vovk et~al.(2005)Vovk, Gammerman, and Shafer]{vovk2005algorithmic}
Vovk, V., Gammerman, A., and Shafer, G.
\newblock \emph{Algorithmic learning in a random world}, volume~29.
\newblock Springer, 2005.

\bibitem[Wang et~al.(2020)Wang, Liu, Liang, Joshi, and Poor]{wang2020tackling}
Wang, J., Liu, Q., Liang, H., Joshi, G., and Poor, H.~V.
\newblock Tackling the objective inconsistency problem in heterogeneous federated optimization.
\newblock \emph{Advances in neural information processing systems}, 33:\penalty0 7611--7623, 2020.

\bibitem[Wang et~al.(2021)Wang, Xu, Wang, and Zhu]{wang2021addressing}
Wang, L., Xu, S., Wang, X., and Zhu, Q.
\newblock Addressing class imbalance in federated learning.
\newblock In \emph{Proceedings of the AAAI Conference on Artificial Intelligence}, volume~35, pp.\  10165--10173, 2021.

\bibitem[Wei et~al.(2020)Wei, Li, Ding, Ma, Yang, Farokhi, Jin, Quek, and Poor]{wei2020federated}
Wei, K., Li, J., Ding, M., Ma, C., Yang, H.~H., Farokhi, F., Jin, S., Quek, T.~Q., and Poor, H.~V.
\newblock Federated learning with differential privacy: Algorithms and performance analysis.
\newblock \emph{IEEE Transactions on Information Forensics and Security}, 15:\penalty0 3454--3469, 2020.

\bibitem[Wu \& Gales(2021)Wu and Gales]{wu2021should}
Wu, X. and Gales, M.
\newblock Should ensemble members be calibrated?
\newblock \emph{arXiv preprint arXiv:2101.05397}, 2021.

\bibitem[Yu et~al.(2022)Yu, Bates, Ma, and Jordan]{yu2022robust}
Yu, Y., Bates, S., Ma, Y., and Jordan, M.
\newblock Robust calibration with multi-domain temperature scaling.
\newblock \emph{Advances in Neural Information Processing Systems}, 35:\penalty0 27510--27523, 2022.

\bibitem[Yurochkin et~al.(2019)Yurochkin, Agarwal, Ghosh, Greenewald, Hoang, and Khazaeni]{yurochkin2019bayesian}
Yurochkin, M., Agarwal, M., Ghosh, S., Greenewald, K., Hoang, N., and Khazaeni, Y.
\newblock Bayesian nonparametric federated learning of neural networks.
\newblock In \emph{International conference on machine learning}, pp.\  7252--7261. PMLR, 2019.

\bibitem[Zadrozny \& Elkan(2001)Zadrozny and Elkan]{zadrozny2001obtaining}
Zadrozny, B. and Elkan, C.
\newblock Obtaining calibrated probability estimates from decision trees and naive bayesian classifiers.
\newblock In \emph{Icml}, volume~1, pp.\  609--616, 2001.

\bibitem[Zadrozny \& Elkan(2002)Zadrozny and Elkan]{zadrozny2002transforming}
Zadrozny, B. and Elkan, C.
\newblock Transforming classifier scores into accurate multiclass probability estimates.
\newblock In \emph{Proceedings of the eighth ACM SIGKDD international conference on Knowledge discovery and data mining}, pp.\  694--699, 2002.

\bibitem[Zhang et~al.(2020)Zhang, Li, Xia, Wang, Yan, and Liu]{zhang2020batchcrypt}
Zhang, C., Li, S., Xia, J., Wang, W., Yan, F., and Liu, Y.
\newblock $\{$BatchCrypt$\}$: Efficient homomorphic encryption for $\{$Cross-Silo$\}$ federated learning.
\newblock In \emph{2020 USENIX annual technical conference (USENIX ATC 20)}, pp.\  493--506, 2020.

\bibitem[Zhang et~al.(2022{\natexlab{a}})Zhang, Li, Li, Xu, Wu, Ding, and Wu]{fl-calibration-zhang2022federated}
Zhang, J., Li, Z., Li, B., Xu, J., Wu, S., Ding, S., and Wu, C.
\newblock Federated learning with label distribution skew via logits calibration.
\newblock In \emph{International Conference on Machine Learning}, pp.\  26311--26329. PMLR, 2022{\natexlab{a}}.

\bibitem[Zhang et~al.(2022{\natexlab{b}})Zhang, Li, Li, Xu, Wu, Ding, and Wu]{zhang2022federated}
Zhang, J., Li, Z., Li, B., Xu, J., Wu, S., Ding, S., and Wu, C.
\newblock Federated learning with label distribution skew via logits calibration.
\newblock In \emph{International Conference on Machine Learning}, pp.\  26311--26329. PMLR, 2022{\natexlab{b}}.

\bibitem[Zhao et~al.(2021)Zhao, Kim, Sahoo, Ma, and Ermon]{zhao2021calibrating}
Zhao, S., Kim, M., Sahoo, R., Ma, T., and Ermon, S.
\newblock Calibrating predictions to decisions: A novel approach to multi-class calibration.
\newblock \emph{Advances in Neural Information Processing Systems}, 34:\penalty0 22313--22324, 2021.

\bibitem[Zhao et~al.(2018)Zhao, Li, Lai, Suda, Civin, and Chandra]{zhao2018federated}
Zhao, Y., Li, M., Lai, L., Suda, N., Civin, D., and Chandra, V.
\newblock Federated learning with non-iid data.
\newblock \emph{arXiv preprint arXiv:1806.00582}, 2018.

\end{thebibliography}
\bibliographystyle{icml2024}

\newpage
\appendix
\numberwithin{equation}{section}
\setcounter{equation}{0}
\renewcommand{\theequation}{\thesection.\arabic{equation}}
\onecolumn

\section{Proof of Theorem \ref{th:global_calibration error}}\label{proof}
\numberwithin{equation}{section}
\setcounter{equation}{0}
\renewcommand{\theequation}{\thesection.\arabic{equation}}

\begin{proof}
    First of all, we show that if $f^*$ is the risk minimizer of NLL loss over $\prob(x, y)$, it also minimizes the calibration error.
    if $f^*$ is a risk minimizer of NLL, $f^*(x)  = p(y=1|x)$. Then $\expectation[y|f^*(x)] = f^*(x)$, thus $f^*$ also minimize the the calibration error as $\text{CE}(f^*)=0$.In binary classification, we can rewrite  the calibration error as 
    \begin{equation}
    \begin{aligned}
        \text{CE}(f) &= \left( \mathbb{E}_{\prob(x, y)}\left[\left|\mathbb{E}[y|f(x)]-f(x)\right|^2\right] \right)^{\frac{1}{2}}\\
        &= \mathbb{E}_{P(x)}\left[P(y=1|x) \cdot (1 - f(x))^2 + P(y=0|x) \cdot (f(x))^2\right].
    \end{aligned}
    \end{equation}
    This equation relates the calibration error $\text{CE}(f^n) -\text{CE}(f^*) = \text{CE}(f^n)$ with $f^R(\vect{x}) - f^*(\vect{x})$.

    \begin{theorem}\label{th:bounded graident dissmilarity} (Adopted from \citep{karimireddy2020scaffold}.) When the gradients of the local loss function and global function have bounded gradient dissimilarity, stated as 
\begin{equation}
    \frac{1}{N}\sum_{i=1}^{N}\lVert \nabla \textnormal{NLL}_c (f(\boldsymbol{x}_i), y_i)\rVert^{2}\leq C_1^{2}+C_2^{2}\lVert\nabla \textnormal{NLL}(f(\boldsymbol{x}_i), y_i)\rVert^{2}\:,\:\forall\boldsymbol{x}_i, y_i\:.
\end{equation} where $C_1$ and $C_2$ are constants s.t. $C_1 \geq 0$ and $C_2\geq 1$, there exists $\mu$-convex function for which FedAvg with more than two clients has an error 
\begin{equation}
    f^R(\vect{x})-f^*(\vect{x})\geq\Omega\biggl(\frac{C_1^{2}}{\mu R^{2}}\biggr).
\end{equation}
\end{theorem}
To leverage Theorem $\ref{th:bounded graident dissmilarity}$ to bound the deviation of $f^n(x) - f^*(x)$ , denote the $\nabla \textnormal{NLL}(f(\vect{x}, y))$ as $h(\vect{x}, y)$, we are required to show that 
\begin{equation}\label{eq:1}
    \sum_{i=1}^{C}w_i\expectation_{\prob_i(\vect{x}, y)}[h(\vect{x}, y)] \leq C_2^2 \expectation_{\prob(\vect{x}, y)}[h(\vect{x}, y)] + C_1^2.
\end{equation}
According to the Definition \ref{def:label_skew} of label skew, we can rewrite equation \ref{eq:1} as 
\begin{equation}
    \sum_{i=1}^{C}w_i\expectation_{\prob_i(y)}\expectation_{\prob(\vect{x}|y)}[h(\vect{x}, y)] \leq C_2^2\expectation_{\prob(y)}\expectation_{\prob(\vect{x}|y)}[h(\vect{x}, y)] + C_1^2.
\end{equation}
Since we assume $\prob(\vect{x}|y)$ is the same across clients, denote $\expectation_{\prob(x|y)}[h(\vect{x}, y)]$ as $g(\vect{x})$, our objective is to show that 
\begin{equation}
    \sum_{i=1}^{C}w_i\expectation_{\prob_i(y)}[g(\vect{x}, y)] \leq C_2^2\expectation_{\prob(y)}[h(x, y)] + C_1^2.
\end{equation}
According to Assumption \ref{ass:distribution_divergence}, we have 
\begin{equation}
    G \coloneqq \sup_i \text{D}_{KL}(\prob_i(y), \prob(y)).
\end{equation}
Suppose $g(\vect{x}, y)$ is bounded by $M$, which we will verify later, by total variation distance and Pinsker's Inequality, we have
\begin{equation}
|\expectation_{\prob_i(y)}[g(\vect{x}, y)] - \expectation_{\prob(y)}[g(\vect{x}, y)]| \leq M \cdot D_{TV}(\prob_i(y), \prob(y)) \leq M \sqrt{\frac{1}{2}G},
\end{equation} where $D_{TV}$ is the total variation distance.
Since, $w_i \geq 0$ and $\sum_{i=1}^{K}w_i=1$, due to the linearity of expectation and triangular inequality, we have 
\begin{equation}\label{eq:2}
    |\sum_{i=1}^{C}w_i\expectation_{\prob_i(y)}[g(\vect{x}, y)] - \expectation_{\prob(y)}[g(\vect{x}, y)]| \leq \sum_{i=1}^{C}w_i |\expectation_{\prob_i(y)}[g(\vect{x}, y)]  - \expectation_{\prob(y)}[g(\vect{x}, y)]|.
\end{equation}
and due to $\sum_{i=1}^{C}w_i$, the $RHS$ of equation \ref{eq:2} 
\begin{equation}
    RHS \leq \sum_{i=1}^Cw_i\cdot M\sqrt{\frac{1}{2}G} = M\sqrt{\frac{1}{2}G}.
\end{equation}
By adding $\expectation_{\prob(y)}[g(\vect{x}, y)]|$ on $LHS$ of equation \ref{eq:2}, we have
\begin{equation}
LHS + |\expectation_{\prob(y)}[g(\vect{x}, y)]| \leq M\sqrt{\frac{1}{2}G}+ |\expectation_{\prob(y)}[g(\vect{x}, y)]|,
\end{equation}
and again, by triangular inequality, we have
\begin{equation}
    LHS + |\expectation_{\prob(y)}[g(\vect{x}, y)]| \geq |LHS + \expectation_{\prob(y)}[g(\vect{x}, y)]| =|\sum_{i=1}^{C}w_i\expectation_{\prob_i(y)}[g(\vect{x}, y)]|.
\end{equation}
And since $g(\vect{x}, y)$ is the norm of gradient which is positive, we have
\begin{equation}\label{eq:3}
    \sum_{i=1}^{C}w_i\expectation_{\prob_i(y)}[g(\vect{x}, y)] \leq\expectation_{\prob(y)}[g(\vect{x}, y)] + M\sqrt{\frac{1}{2}G}.
\end{equation}
which meets our object with $C_1^2 = M\sqrt{\frac{1}{2}G}$ and $C_2^2 = 1$. Now, let's verify that  $g(\vect{x}, y)$ is indeed bounded by a constant $M$. Here, we make one additional assumption that 
\begin{assumption}\label{ass:feature}
the extracted feature $z \coloneqq \vect{\theta}(\vect{x})$ of samples in the same class are similar. To be more specific, $\text{Var}_{\prob(\vect{x}|y)}(z) \leq S$ where $S$. This assumption is empirically verified in \cite{wang2021addressing} and also used in \cite{fl-calibration-zhang2022federated}. 
\end{assumption}
This assumption implies that gradient difference among clients only happens at the last layer. So that we can explicitly write $g(\vect{x}, y)$ as 
\begin{equation}
\begin{aligned}
   g(\vect{x}, y) =     \expectation_{\prob(\vect{x}|y)}[|\nabla \text{NLL}(\vect{x}, y)|^2] = 
\expectation_{\prob(\vect{x}|y)}[|(\sigma(w^\top z + b)-y)*z|^2].
\end{aligned}
\end{equation}
Note that the existence of variance in Assumption \ref{ass:feature} implies the existence of the mean $\overline{z}$. Now, we have
\begin{equation}
    \expectation_{{\prob(\vect{x}|y)}}[|(\sigma(w^\top z + b)-y)*z|^2] \leq \expectation_{{\prob(\vect{x}|y)}}[|z|^2] = \text{var}^2(z) + \overline{z}^2 \leq S^2 + \overline{z}^2.
\end{equation}
Now replacing $M$ in\ref{eq:3} with $(\overline{z}^2+ S^2)$, we have
\begin{equation}
        \sum_{i=1}^{C}w_i\expectation_{\prob_i(y)}[g(\vect{x}, y)] \leq\expectation_{\prob(y)}[g(\vect{x}, y)] + (\overline{z}^2+ S^2)\sqrt{\frac{1}{2}G}.
\end{equation}
By Theorem \ref{th:bounded graident dissmilarity}, we have
\begin{equation}
    f^R(\vect{x})-f^*(\vect{x})\geq\Omega\bigg(\frac{(\overline{z}^2+ S^2)\sqrt{\frac{1}{2}G }}{\mu\boldsymbol{R}^{2}}\bigg).
\end{equation}
Lastly, we investigate how this deviation propagates through the calibration error
\begin{equation}
\begin{aligned}
    \overline{\text{CE}}(f^n) &= \Big(\expectation_{\prob(\vect{x}, y)}[|\prob(y=1|\vect{x})-f^R(\vect{x})|^2]\Big)^{\frac{1}{2}} \\
 &=  \Big(\expectation_{\prob(\vect{x}, y)}[|f^*(\vect{x}) - f^R(\vect{x})|^2]\Big)^{\frac{1}{2}}\\
&\geq \expectation_{\prob(x, y)}[|f^*(\vect{x}) - f^R(\vect{x})|]   &\text{Jensen's Inequality} \\
& \geq \Omega\bigg(\frac{(\overline{z}^2+ S^2)\sqrt{\frac{1}{2}G }}{\mu\boldsymbol{R}^{2}}\bigg).
\end{aligned}
\end{equation}
\end{proof}

\section{Examples of Order Preserving Network}\label{example_op}
Theorem 1 in \citep{rahimi2020intra} states that a continuous function $f: \mathbb{R}^n \mapsto \mathbb{R}^n$ is order-preserving, if and only if $f(\mathbf{x}) = S^{-1}(\mathbf{x})U\mathbf{w}(\mathbf{x})$ where $U$ is an upper-triangular matrix of ones and 
$\mathbf{w}: \mathbb{R}^n \mapsto \mathbb{R}^n$ s.t. $\begin{cases}\mathbf{w}_i(\mathbf{x}) = 0, \mathbf{y}_i = \mathbf{y}_{i+1}  \text{ and } i < n\\\mathbf{w}_i(\mathbf{x}) > 0, \mathbf{y}_i > \mathbf{y}_{i+1} \text{ and } i < n \\\mathbf{w}_n(\mathbf{x}) \text{ is arbitrary }\end{cases}$
where $\mathbf{y} = S(\mathbf{x})\mathbf{x}$ is the sorted version of $\mathbf{x}$. 

The order-preserving network is a direct application of this theorem, where $S(\mathbf{x})$ is achieved using the sorting component and the element-wise production between activation after softplus and the $\mathbf{y}_i - \mathbf{y}_{i+1}$ ensures $\mathbf{w}_i(\mathbf{x})  \geq 0$ when $\mathbf{y}_i - \mathbf{y}_{i+1} \geq 0$. To verify this theorem, consider an unsorted input vector $\mathbf{x} = [3, 4, 2, 2]^\top$, where the correct order should be $\mathbf{x}_2 > \mathbf{x}_1  > \mathbf{x}_3 = \mathbf{x}_4$. In other words, $\mathbf{y}=S(\mathbf{x})\mathbf{x} = [\mathbf{x}_2, \mathbf{x}_1, \mathbf{x}_3, \mathbf{x}_4]^\top$ if we prefer $\mathbf{x}_3$ or $[\mathbf{x}_2, \mathbf{x}_1, \mathbf{x}_4, \mathbf{x}_3]^\top$ otherwise.
Then, $\mathbf{w} = [\mathbf{w}_1>0, \mathbf{w}_2>0, \mathbf{w}_3=0, \mathbf{w}_4]^\top$. 
$U\mathbf{w} = \begin{bmatrix} \mathbf{w}_1 + \mathbf{w}_2 + \mathbf{w}_3 + \mathbf{w}_4 \\ \mathbf{w}_2 + \mathbf{w}_3 + \mathbf{w}_4\\ \mathbf{w}_3 + \mathbf{w}_4 \\ \mathbf{w}_4\end{bmatrix}$ is monotonically non-decreasing. 
Once we permute $U\mathbf{w}$ back by exchanging the index 1 and 2, $f(\mathbf{x}) = S^{-1}(x)U\mathbf{w}$ also has the ordering $f(\mathbf{x})_2 > f(\mathbf{x})_1  > f(\mathbf{x})_3 = f(\mathbf{x})_4$.

\section{Discussion on Baselines}\label{Ap:basline}

As part of our experimental design, we benchmark our framework against intuitive scaler designs and aggregation strategies, and highlight their correspondence in existing works.

The simplest aggregation strategy involves disregarding the parameterization of scalers and adopting a direct averaging of the model's scaled probability, which is equivalent to Deep Ensemble \cite{lakshminarayanan2017simple, rahaman2021uncertainty} and
We denote the methods as \textsc{Ens.}.Within our framework, the ensemble method is expressed as 
\begin{equation}    
\overline{\phi}\circ f^R(\vect{x}_i) = \frac{1}{C}\sum_{c=1}^{C}\phi_c\circ f^R(\vect{x}_i). 
\end{equation}
Despite the simplicity of \textsc{Ens}, it has been demonstrated to yield robust uncertainty quantification, often outperforming more elaborate methods \citep{lakshminarayanan2017simple, lee2015m}. 

An intuitive extension to \textsc{Ens} is to average the temperature parameter of individual temperature scalers to
\begin{equation}
   \overline{\phi}\circ f^R(\vect{x}_i) = \sigma\bigg(\frac{\vect{\theta}^R(\vect{x}_i)}{\sum_{c=1}^{C}T_c}\bigg) 
\end{equation} referred to as\textsc{AvgT}. \textsc{AvgT} and \textsc{Ens} share similar advantages and limitations. Both methods can be applied post-training, negating the need for communication overhead However, \citet{rahaman2021uncertainty} reveal that \textsc{Ens}  only works when $\phi_c\circ f^R(\vect{x}_i)$ is overconfident and it does not meaningfully contribute to an ensemble’s uncertainty quantification when $\prob_c(x, y)$ is different from $\prob(x, y)$ \citep{abe2022deep}. The simplicity of these methods limits their generalization capabilities.

MD-TS \citep{yu2022robust} proposes to use a linear regression model to  regression estimate the temperature 
$T$ from the last layer's output $\vect{\theta}_i(x)$. This approach improves the model's calibration under distribution shifts. MD-TS aligns with our analysis as linear regressor offers better generalization capability and scaling through temperature does not affect accuracy. We adapt this concept for federated settings. Unlike the original approach, which requires data sharing and a validation dataset $\mathcal{D}_{val}$, we suggest leveraging FedAvg for deriving the linear model, bypassing the need for direct access to $\mathcal{D}_{val}$, we denote this method as \textsc{LR-TS}.

\textsc{LR-TS} can be formulated as:
A client maintains a linear regression mode $lr_c$ with the aim of predicting the $T_c$. 

$lr_c(\vect{\theta}\vect{x}_i; W, b)$ is trained using 
\begin{equation}
    lr_c^* = \arg_{lr} \min\sum_{i=1}^{|\mathcal{D}_{val}|}(W\vect{\theta}^R\vect{x}_i + b)-T_c
\end{equation}
Then the global calibration with scaler $\overline{\phi}$ can be formulated as
\begin{equation}
\overline{\phi}\circ f^R(\vect{x}_i) = \sigma\bigg(\frac{\vect{\theta}^R(\vect{x}_i)}{lr(\vect{\theta}^R(\vect{x}_i))}\bigg)
\end{equation}
, where $lr\coloneqq\text{FedAvg}(\{lr_c\})$.

\section{The Weight Matching Algorithm}\label{Ap:wm}
For ease of reference, we list the Weight Matching algorithm from \cite{ainsworth2022git} here.

\begin{algorithm}
\caption{WeightMatching \cite{ainsworth2022git}}
\label{alg:WeighMatching}
\begin{algorithmic}[1]
\STATE \textbf{procedure} $\text{WeightMatching}(\overline{\vect{\pi}}^0,
    \vect{\pi}_c)$
    \STATE \textbf{Given:} 
    \STATE
    Global scaler weights $\overline{\vect{\pi}}^0 = \{W_1^{(A)}, \ldots, W_L^{(A)}\}$ and local scaler weight $\vect{\pi}_c = \{W_1^{(B)}, \ldots, W_L^{(B)}\}$
    \STATE \textbf{Result:} A permutation $\mathbf{M} = \{P_1, \ldots, P_{L-1}\}$ of $\vect{\pi}_c$ such that $\text{vec}(\overline{\vect{\pi}}^0) \cdot \text{vec}(\pi(\vect{\pi}_c))$ is approximately maximized.
    \STATE \textbf{Initialize:} $P_1 \leftarrow I, \ldots, P_{L-1} \leftarrow I$
    \REPEAT
        \FOR{$\ell$ in RANDOMPERMUTATION$(1, \ldots, L - 1)$}
            \STATE $P_{\ell} \leftarrow \text{SOLVELAP}(W_{\ell}^{(A)} P_{\ell-1} (W_{\ell}^{(B)})^T + (W_{\ell+1}^{(A)})^T P_{\ell+1} W_{\ell+1}^{(B)})$
        \ENDFOR
    \UNTIL{convergence}
    \STATE
    \textbf{return} $\mathbf{M}$
\end{algorithmic}
\end{algorithm}

\end{document}